\documentclass{article} %
\usepackage{iclr2027_conference,times}

\usepackage{amsmath,amsfonts,bm}

\def\eqref#1{equation~\ref{#1}}

\def\1{\bm{1}}

\DeclareMathAlphabet{\mathsfit}{\encodingdefault}{\sfdefault}{m}{sl}
\SetMathAlphabet{\mathsfit}{bold}{\encodingdefault}{\sfdefault}{bx}{n}

\usepackage{hyperref}
\usepackage{fontawesome5}
\usepackage{url}
\usepackage{algorithm}
\usepackage{algorithmic}
\usepackage{xcolor}
\usepackage{amsmath}
\usepackage{amssymb}
\usepackage{amsthm}
\usepackage{booktabs}
\usepackage{tabularx}
\usepackage{array}
\usepackage{multirow}
\usepackage[table]{xcolor}
\usepackage{colortbl}
\usepackage{makecell}
\usepackage{graphicx}
\usepackage{enumitem}
\usepackage{listings}
\usepackage{titletoc}

\definecolor{mygray}{RGB}{240,240,240}      %
\definecolor{myblue}{RGB}{230,240,255}
\definecolor{mygreen}{RGB}{240,255,240}

\title{EvoSCM: Scientific Belief Revision Through Causal Model Evolution and Experimentation}

\author{%
\makebox[\linewidth][c]{\bfseries
  Qing Zhao\textsuperscript{1} \quad
  Haowei Li\textsuperscript{1} \quad
  Weijian Deng\textsuperscript{2} \quad
  Sibei Yang\textsuperscript{1} \quad
  Pengxu Wei\textsuperscript{1} \quad
  Liang Lin\textsuperscript{1}}\\[3pt]
\makebox[\linewidth][c]{%
  \textsuperscript{1}Sun Yat-sen University}\\[-1pt]
\makebox[\linewidth][c]{%
  \textsuperscript{2}Tsinghua Shenzhen International Graduate School, Tsinghua University}\\[3pt]
\makebox[\linewidth][c]{\ttfamily\small
  \{zhaoq78, lihw59\}@mail2.sysu.edu.cn \quad
  dengwj16@sz.tsinghua.edu.cn}\\[-1pt]
\makebox[\linewidth][c]{\ttfamily\small
  yangsb3@mail.sysu.edu.cn \quad
  weipx3@mail.sysu.edu.cn \quad
  linliang@ieee.org}
}

\iclrfinalcopy %
\begin{document}

\maketitle
\lhead{Preprint}

\begin{abstract}
Scientific discovery depends on the ability to form hypotheses, test them through experiments, and revise them when evidence disagrees. Existing LLM agents support this process by improving their reasoning or actions, but their scientific beliefs are often scattered across free-form reasoning and difficult to update coherently. This makes it difficult to identify what failed, what should change, and whether revisions remain consistent with prior evidence. We introduce EvoSCM, which represents scientific beliefs as a population of structural causal model (SCM) hypotheses that can be tested and revised across experiments. EvoSCM formulates scientific discovery as a closed loop in which causal hypotheses guide experimentation and experimental outcomes drive causal model evolution.
Competing SCM hypotheses make falsifiable predictions and guide discriminative experiments that separate alternative explanations. When observations contradict these predictions, EvoSCM distills discrepancies into correction rules identifying which aspects of the hypotheses fail to explain the evidence. These rules guide revisions to causal dependencies, latent factors, mechanisms, and parameters. Revised hypotheses are validated against accumulated evidence and carried forward to guide subsequent experiments, allowing scientific beliefs to evolve cumulatively.
We evaluate EvoSCM across physics, chemistry and materials, and biology. It consistently outperforms baseline agents and existing evolution methods, yielding more accurate explanations and predictions with more effective use of experimental budgets. The evolved SCMs also transfer across base models, suggesting reusable scientific knowledge beyond any single model's reasoning process.
\end{abstract}

\begin{center}
\small
\faLink\ Project Page: \href{https://evoscm.github.io/}{\texttt{evoscm.github.io}} \quad \faGithub\ Code: \href{https://github.com/evoscm/EvoSCM}{\texttt{github.com/evoscm/EvoSCM}}
\end{center}

\section{Introduction}

Scientific discovery is inherently iterative: hypotheses are proposed, tested through experiments, and revised as new evidence accumulates. Recent agents can improve how they reason and act through prompts, memories, skills, or workflows~\citep{gao2026survey,agrawal2026gepa,ouyang2026reasoningbank,yang2026skillopt,zhang2025aflow,nguyen2026recursive}, and can increasingly generate scientific hypotheses, design experiments, and interpret outcomes~\citep{boiko2023autonomous,lu2024ai,huang2025automated,abhyankar2026llema,kabra2026llm,wiemann2026discoverphysics}. These capabilities improve how agents carry out scientific tasks, but they do not necessarily provide an explicit and revisable representation of what the agent currently believes about the underlying system. Instead, such beliefs often remain distributed across free-form reasoning traces, experimental records, or memory summaries~\citep{takahara2026toward}. This makes it difficult to connect hypotheses with the evidence that supports, contradicts, or motivates their revision.

This limitation becomes particularly important when new evidence contradicts a prediction. The agent must determine which assumption may be responsible, what part of its current explanation should be revised, and whether the revised explanation remains consistent with prior observations. When scientific beliefs are not represented explicitly, it becomes difficult to trace prediction failures back to specific assumptions and to propagate the resulting revisions coherently across subsequent experiments. Thus, scientific agents can struggle to revise their hypotheses reliably when confronted with contradictory evidence~\citep{rios2026ai}. This motivates an explicit representation of scientific beliefs that links hypotheses, predictions, and experimental evidence, and can be tested and revised as new evidence arrives.

\begin{figure}
    \centering
    \includegraphics[width=\linewidth]{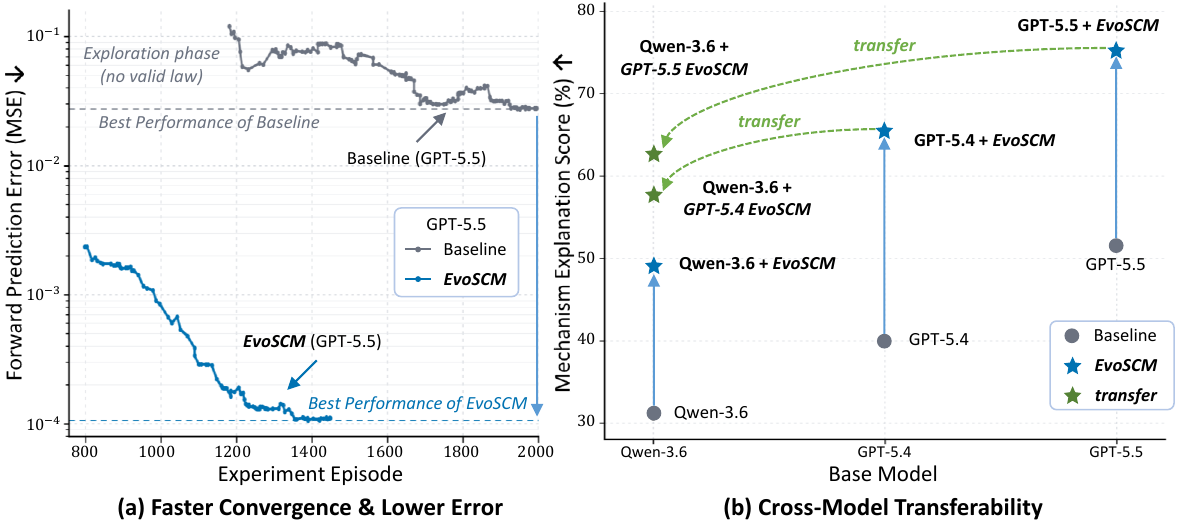}
    \caption{\textbf{Discovery performance and transferability of EvoSCM on DiscoverPhysics}~\citep{wiemann2026discoverphysics}. (a) \emph{Faster Convergence \& Lower Error}: EvoSCM achieves lower prediction error while requiring fewer experimental episodes than the baseline agent. (b) \emph{Cross-Model Transferability}: evolved SCMs can be transferred directly across different base models.}
    \label{fig:first_image}
\end{figure}

To this end, we introduce EvoSCM, a framework that represents the agent's evolving scientific beliefs as a population of competing structural causal model (SCM) hypotheses. Each hypothesis provides an explicit causal explanation through its graph structure, latent factors, functional mechanisms, and parameters, making the agent's current understanding directly testable and revisable. EvoSCM couples this representation with experimentation in a closed loop. In each round, the current hypotheses abductively interpret accumulated evidence, guide interventions that discriminate among competing explanations, and make falsifiable predictions. Prediction--observation discrepancies are inductively distilled into correction rules that drive targeted revisions to causal structure, latent variables, mechanisms, and parameters. The revised hypotheses are then deductively evaluated against accumulated evidence and structural consistency before guiding the next round of experimentation.
In this way, EvoSCM extends agent evolution from improving how an agent reasons to also revising what the agent believes about the world.

We evaluate EvoSCM across three scientific domains: non-canonical physical law discovery (DiscoverPhysics~\citep{wiemann2026discoverphysics}), chemistry and materials discovery (LLEMA~\citep{abhyankar2026llema}), and biological network inference (ActiveSciBench-GRN~\citep{kabra2026llm}). Across all three domains, EvoSCM consistently outperforms baseline agents and existing evolution methods, yielding more accurate explanations and predictions while using experimental budgets more effectively (Figure~\ref{fig:first_image}(a)). Moreover, the evolved SCMs transfer across base architectures, suggesting that the learned causal models capture reusable scientific knowledge beyond any particular model's reasoning process (Figure~\ref{fig:first_image}(b)).

\section{Related Work}
\label{sec:related_work}

\noindent\textbf{Self-Evolving LLM Agents.}
LLM agents can improve from experience without updating their parameters~\citep{gao2026survey}, for example by optimizing prompts~\citep{agrawal2026gepa}, distilling reasoning trajectories into reusable memories~\citep{ouyang2026reasoningbank}, acquiring skills~\citep{wang2023voyager,yang2026skillopt}, searching over workflows~\citep{zhang2025aflow}, or recursively refining agent states~\citep{nguyen2026recursive}. A common thread across these approaches is that experience improves the agent's \emph{procedure}: how it reasons, plans, and acts. EvoSCM targets a complementary axis, using experience to revise the agent's \emph{epistemic state}: its explicit model of how the external world works.

\noindent\textbf{Scientific Agents.}
Recent systems automate parts of the scientific pipeline~\citep{boiko2023autonomous,lu2024ai,huang2025automated}, yet evidence-driven belief revision could be fragile: refutation rarely triggers genuine updates to an agent's hypotheses~\citep{rios2026ai}. Among hypothesis-driven discovery systems, LLM-AutoSciLab~\citep{kabra2026llm} evolves competing hypotheses and selects discriminative experiments, LLEMA~\citep{abhyankar2026llema} applies evolutionary search to multi-objective materials discovery, and PiEvo~\citep{pu2026principle} evolves natural-language scientific principles via uncertainty minimization.
In contrast, EvoSCM combines executable causal models with structural revisability, grounding belief revision in falsifiable, mechanism-level predictions.

\section{EvoSCM: Causal Model Evolution}
\label{sec:method}

\begin{figure}
    \centering
    \includegraphics[width=\linewidth]{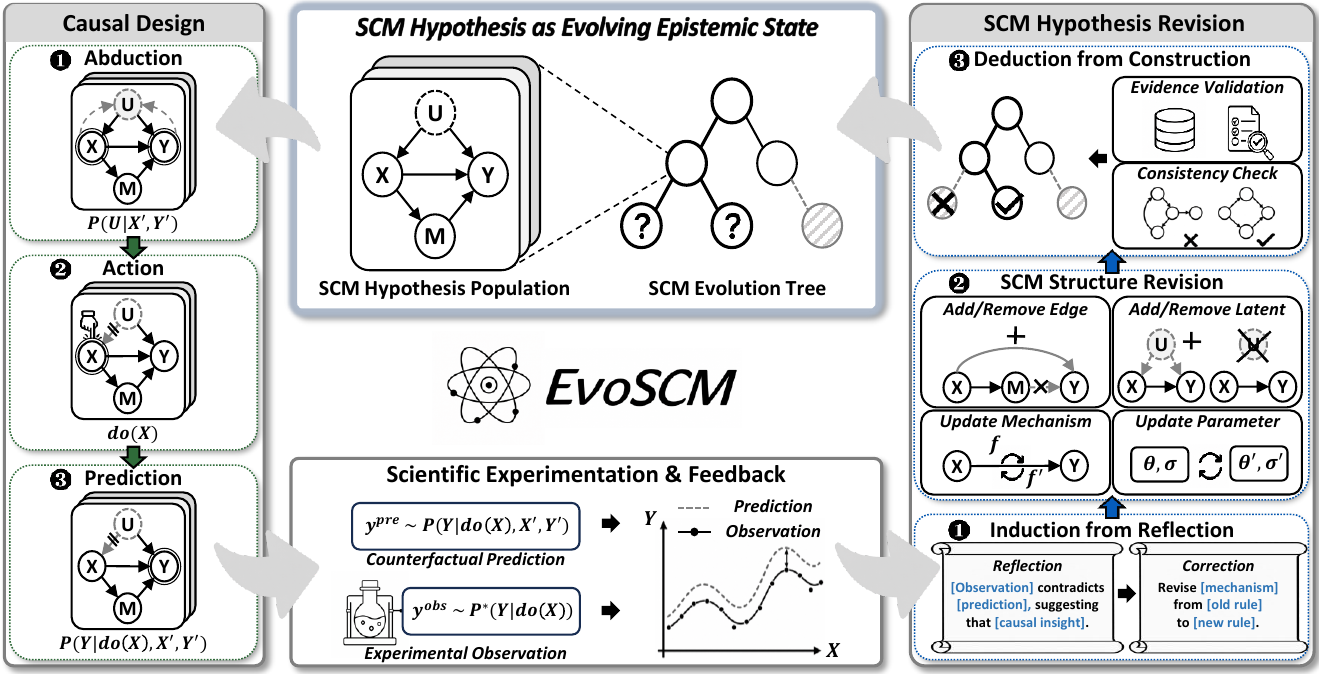}
\caption{\textbf{Overview of EvoSCM.} EvoSCM maintains a population of competing SCM hypotheses and evolves it through a closed discovery loop: (1) \emph{Causal Experiment Design}: the agent abductively interprets accumulated evidence, selects discriminative interventions, and commits to falsifiable predictions tested through experimentation; (2) \emph{SCM Hypothesis Revision}: prediction--observation discrepancies are inductively distilled into correction rules that guide targeted revisions to causal structure and mechanisms, while revised hypotheses are deductively validated against accumulated evidence and structural consistency to form the next-round population.}
    \label{fig:method}
\end{figure}

\subsection{SCM Hypothesis as Evolving Epistemic State}
\label{sec:epistemic_state}

\noindent\textbf{Scientific Discovery Formulation.}
Scientific discovery is the process of uncovering the laws and principles that govern observed phenomena through iterative hypothesis formation, experimental testing, and belief revision~\citep{popper2005logic}.
We formalize agentic scientific discovery as the sequential recovery of an unknown causal model through experimentation. An agent interacts with an environment~$\mathcal{E}$ governed by a true structural causal model $\mathcal{M}^* = (\mathbf{V},\, \mathbf{U}^*,\, G^*,\,\mathbf{F}^*,\, \boldsymbol{\Theta}^*)$~\citep{pearl2009causality}, where $\mathbf{V} = \{V_i\}_{i=1}^{n}$ are observable endogenous variables, $\mathbf{U}^* = \{U_i^*\}_{i=1}^{n}$ are mutually independent latent exogenous variables, $G^*$ is a directed acyclic graph (DAG) encoding the true causal structure, $\mathbf{F}^* = \{f_i^*\}_{i=1}^{n}$ are the structural equations (causal mechanisms) with $V_i = f_i^*(\mathrm{Pa}_i^{G^*},\, U_i^*;\, \Theta_i^*)$, and $\boldsymbol{\Theta}^* = \{\Theta_i^*\}_{i=1}^{n}$ parameterizes these mechanisms.
Discovery proceeds over a budget of $T$ rounds. At each round~$t$, the agent selects an intervention $\mathrm{do}(\mathbf{X}_t = \mathbf{x}_t)$ on a subset $\mathbf{X}_t \subseteq \mathbf{V}$ and observes outcomes $\mathbf{y}_t \sim P^*(\mathbf{Y}_t \mid \mathrm{do}(\mathbf{X}_t\!=\!\mathbf{x}_t))$ for the remaining variables $\mathbf{Y}_t = \mathbf{V} \setminus \mathbf{X}_t$. The accumulated evidence through round~$t$ is $\mathcal{D}_t = \{(\mathrm{do}(\mathbf{x}_\tau),\,\mathbf{y}_\tau)\}_{\tau=1}^{t}$.

The agent models the environment with an SCM hypothesis $\mathcal{H} = (\mathbf{V},\, \mathbf{U},\, G,\, \mathbf{F},\, \boldsymbol{\Theta})$ that shares the observable variables~$\mathbf{V}$ but posits its own latent variables, causal graph, mechanisms, and parameters. The goal of agentic scientific discovery is to evolve~$\mathcal{H}$ over the $T$ rounds so that it recovers the causal structure, mechanisms, and predictive behavior of~$\mathcal{M}^*$. An SCM is a natural representation for the epistemic state of a scientific agent: its causal graph~$G$ encodes qualitative causal structure, while its structural equations~$\mathbf{F}$ with parameters~$\boldsymbol{\Theta}$ encode quantitative mechanisms. Together, $(G, \mathbf{F}, \boldsymbol{\Theta})$ constitutes a complete, falsifiable scientific hypothesis: the $\mathrm{do}(\cdot)$ operator~\citep{pearl2009causality} allows the agent to derive interventional predictions that can be directly tested against experimental outcomes, and any prediction--observation discrepancy can be traced to specific structural or parametric commitments within the hypothesis.

\noindent\textbf{Population-Based Epistemic State.}
A single hypothesis, however, may be insufficient for effective scientific discovery. Multiple causal structures can be consistent with the evidence available at any given round, and prematurely committing to one risks overlooking the true mechanism.
EvoSCM therefore maintains a population of competing hypotheses $\mathcal{P}_t = \{\mathcal{H}_1^t, \ldots, \mathcal{H}_K^t\}$ as its epistemic state at round~$t$. Each hypothesis $\mathcal{H}_k^t = (\mathbf{V},\, \mathbf{U}_k^t,\, G_k^t,\, \mathbf{F}_k^t,\, \boldsymbol{\Theta}_k^t)$ encodes a distinct candidate causal explanation of~$\mathcal{E}$. 
Maintaining a diverse population serves two purposes: it guards against premature convergence to an incorrect explanation, and the disagreement among competing hypotheses directly guides experiment design by identifying interventions under which current candidates make maximally divergent predictions~\citep{lindley1956measure,box1967discrimination}.

\noindent\textbf{Overview of EvoSCM.}
As shown in Figure~\ref{fig:method}, EvoSCM maintains a population of competing SCM hypotheses~$\mathcal{P}_t$ as the agent's current scientific belief state and evolves this population through a closed discovery loop. Each round consists of two complementary phases: the current hypothesis population first guides experiment design and prediction, and the resulting experimental feedback then drives hypothesis revision to produce the next population~$\mathcal{P}_{t+1}$.

(1) \textit{Causal Experiment Design.}
Given~$\mathcal{P}_t$ and accumulated evidence, the agent abductively interprets the observations under each hypothesis, selects interventions that discriminate among competing explanations, and commits to falsifiable predictions before experimentation. The experimental outcomes are then compared with these predictions to expose where the current hypotheses fail.

(2) \textit{SCM Hypothesis Revision.}
Prediction--observation discrepancies are inductively distilled into correction rules that guide revisions to causal structure, latent variables, mechanisms, and parameters. The revised hypotheses are used deductively to derive consequences under accumulated interventions and validated against prior evidence and structural consistency. Well-supported hypotheses form the updated population~$\mathcal{P}_{t+1}$, which in turn guides the next round of experimentation.

After $T$~rounds, the evolved SCM population~$\mathcal{P}_T$ encodes the agent's accumulated causal knowledge and can be used for downstream reasoning, prediction, and generalization to unseen interventions.

\subsection{Causal Experiment Design}
\label{sec:experiment_design}

Given the current hypothesis population~$\mathcal{P}_t$ and accumulated evidence~$\mathcal{D}_t$, the agent designs the next experiment by reasoning causally within each hypothesis and comparatively across the population.
This phase follows the logic of Pearl's counterfactual framework~\citep{pearl2009causality}: abduction (infer latent states), action (select an intervention), and prediction (derive expected outcomes), extended with an experimentation step that produces the empirical feedback driving hypothesis revision.

\noindent\textbf{Abduction.}
For each hypothesis~$\mathcal{H}_k^t \in \mathcal{P}_t$, the agent infers the latent exogenous variables most consistent with the accumulated evidence:
\(
  P(\mathbf{U}_k^t \mid \mathcal{D}_t,\, \mathcal{H}_k^t).
\)
This abductive step conditions on all past interventions and observations in~$\mathcal{D}_t$ to determine, under hypothesis~$\mathcal{H}_k^t$, what values of the latent variables~$\mathbf{U}_k^t$ would best explain the observed outcomes.
Because each hypothesis posits its own latent variables and causal structure, the same evidence may yield different abduced latent states across hypotheses, anchoring each candidate in a distinct causal interpretation of the available data.

\noindent\textbf{Action.}
The agent selects the next intervention $\mathrm{do}(\mathbf{X}_{t+1} = \mathbf{x}_{t+1})$ to maximally discriminate among the competing hypotheses in~$\mathcal{P}_t$. Concretely, the agent identifies interventions under which the hypotheses' predicted outcomes diverge most, so that the experimental result will provide maximal evidence for distinguishing between candidate explanations~\citep{lindley1956measure,box1967discrimination}.
This population-level disagreement is what makes experiment design \emph{comparative}: rather than testing a single hypothesis in isolation, the agent exploits the diversity of the population to select experiments whose outcomes are most likely to narrow the set of viable candidates~\citep{chaloner1995bayesian}.

\noindent\textbf{Prediction.}
Before executing the intervention, the agent commits to a falsifiable, hypothesis-specific prediction for each candidate. Under hypothesis~$\mathcal{H}_k^t$ with abduced latent states, the predicted outcome is:
\(
  \mathbf{y}_k^{\mathrm{pre}} \sim
  P\!\left(\mathbf{Y} \mid
  \mathrm{do}(\mathbf{X}_{t+1}\!=\!\mathbf{x}_{t+1}),\,
  \mathcal{D}_t,\, \mathcal{H}_k^t\right).
\)
These predictions are recorded prior to experimentation, ensuring that each hypothesis stakes a concrete, testable claim on the experimental outcome.
This commitment is essential: it prevents post-hoc rationalization and allows subsequent discrepancies to be unambiguously attributed to specific hypotheses.

\noindent\textbf{Experimentation and Feedback.}
The agent executes the selected intervention $\mathrm{do}(\mathbf{X}_{t+1}\!=\!\mathbf{x}_{t+1})$ in the environment~$\mathcal{E}$ and observes the outcome:
\(
  \mathbf{y}_{t+1}^{\mathrm{obs}} \sim
  P^*\!\left(\mathbf{Y} \mid
  \mathrm{do}(\mathbf{X}_{t+1}\!=\!\mathbf{x}_{t+1})\right),
\)
drawn from the true data-generating process~$\mathcal{M}^*$. The agent then compares each hypothesis's committed prediction~$\mathbf{y}_k^{\mathrm{pre}}$ against the observed outcome~$\mathbf{y}_{t+1}^{\mathrm{obs}}$, producing a set of prediction--observation discrepancies that identify where and how each candidate's causal account fails to match reality.
The accumulated evidence is updated as $\mathcal{D}_{t+1} = \mathcal{D}_t \cup \{(\mathrm{do}(\mathbf{x}_{t+1}),\, \mathbf{y}_{t+1}^{\mathrm{obs}})\}$, and both the discrepancies and the updated evidence are passed to the hypothesis revision phase.

\subsection{SCM Hypothesis Revision}
\label{sec:revision}

The prediction--observation discrepancies and updated evidence~$\mathcal{D}_{t+1}$ produced by the experimentation phase drive the evolution of the hypothesis population.
This phase transforms each hypothesis through three steps: induction extracts structured correction rules from empirical discrepancies, revision applies these rules as targeted edits to causal structure and mechanisms, and deduction validates the revised hypotheses against accumulated evidence and structural consistency to produce~$\mathcal{P}_{t+1}$.

\noindent\textbf{Induction from Reflection.}
For each hypothesis~$\mathcal{H}_k^t \in \mathcal{P}_t$, the agent analyzes the discrepancies between its committed predictions~$\mathbf{y}_k^{\mathrm{pre}}$ and the observed outcomes~$\mathbf{y}_{t+1}^{\mathrm{obs}}$ to identify systematic patterns of failure.
Rather than treating each discrepancy as an isolated error, the agent examines the accumulated evidence~$\mathcal{D}_{t+1}$ to detect recurring regularities that the current hypothesis fails to capture. These regularities are distilled into explicit \emph{correction rules}: concise, natural-language descriptions of the empirical pattern that the hypothesis must accommodate.
Correction rules serve as the interface between empirical evidence and structural model editing: they translate quantitative prediction failures into qualitative directives that specify \emph{what} must change in the hypothesis without prescribing \emph{how} that change should be realized at the structural level.

\noindent\textbf{SCM Structure Revision.}
Guided by the correction rules from the induction step, the agent applies targeted revisions to each hypothesis~$\mathcal{H}_k^t$.
EvoSCM supports four revision operators, each modifying a different component of an SCM hypothesis:
(1) \emph{Add\,/\,Remove Edge}: modify the causal graph~$G_k^t$ by inserting a new directed edge to represent a previously unmodeled causal dependency, or removing an existing edge that the evidence no longer supports;
(2) \emph{Add\,/\,Remove Latent}: introduce a new latent variable into~$\mathbf{U}_k^t$ to account for unexplained confounding or mediating effects, or remove a latent variable that has become redundant given the revised structure;
(3) \emph{Update Mechanism}: replace or modify a structural equation $f_i \in \mathbf{F}_k^t$ to change the functional form of a causal mechanism;
(4) \emph{Update Parameter}: adjust the parameters~$\boldsymbol{\Theta}_k^t$ to improve quantitative fit while preserving the current structure and mechanisms.
The agent selects the minimal set of operators sufficient to accommodate the correction rules, preferring shallow revisions (parameter and mechanism updates) when possible and resorting to deeper structural edits (edge and latent changes) only when the evidence warrants them. Each revision produces a candidate revised hypothesis
$\tilde{\mathcal{H}}_k^{t+1} = (\mathbf{V},\, \tilde{\mathbf{U}}_k^{t+1},\, \tilde{G}_k^{t+1},\, \tilde{\mathbf{F}}_k^{t+1},\, \tilde{\boldsymbol{\Theta}}_k^{t+1})$, which is then passed to the deduction step for validation.

\noindent\textbf{Deduction from Construction.}
The deduction step validates each candidate revised hypothesis~$\tilde{\mathcal{H}}_k^{t+1}$ before it is admitted into the next-round population~$\mathcal{P}_{t+1}$. Validation comprises two checks:
(1) \emph{Evidence Validation}: the agent derives predictions from~$\tilde{\mathcal{H}}_k^{t+1}$ for all interventions in the accumulated evidence~$\mathcal{D}_{t+1}$ and verifies that the revised hypothesis accounts for past observations. A hypothesis that resolves the most recent discrepancy but introduces new contradictions with earlier evidence is rejected;
(2) \emph{Consistency Check}: the agent verifies the internal coherence of the revised hypothesis: the causal graph~$\tilde{G}_k^{t+1}$ must remain a valid DAG, the structural equations must be well-defined for all variables given the revised graph, and the posited mechanisms must be mutually compatible.
Hypotheses that pass both checks are retained; those that fail either check are discarded from the population. This selective step instantiates the population-level evolution: individual hypotheses are structurally revised in the preceding step, while deductive validation determines which revised candidates survive into the next round, producing the evolved population~$\mathcal{P}_{t+1}$. The retained population, together with the updated evidence~$\mathcal{D}_{t+1}$, seeds the next round of experiment design.

\subsection{Inference with Evolved SCM}
\label{sec:inference}

After $T$~rounds of the discovery loop, the evolved hypothesis population~$\mathcal{P}_T$ encodes the agent's accumulated causal knowledge as a structured, executable scientific model. Given a novel intervention $\mathrm{do}(\mathbf{X}^{\prime} = \mathbf{x}^{\prime})$ not present in~$\mathcal{D}_T$, each surviving hypothesis derives a prediction through its structural equations, $\mathbf{y}_k^{\prime} \sim P(\mathbf{Y} \mid \mathrm{do}(\mathbf{X}^{\prime}\!=\!\mathbf{x}^{\prime}), \mathcal{D}_T, \mathcal{H}_k^T)$, and when multiple hypotheses survive, predictions are aggregated into a consensus outcome.
This generalization capacity follows from representing knowledge as a causal model: the $\mathrm{do}(\cdot)$ operator derives predictions for unseen interventions by simulating causal consequences through the discovered graph and mechanisms, rather than interpolating from stored input--output associations~\citep{pearl2009causality,bareinboim2016causal}.
The evolved SCM is also decoupled from the experimental trajectory that produced it, constituting a portable, inspectable model that can be transferred to new agents or used to plan future experiments, a property procedural improvements such as better prompts or workflows do not naturally support.

\section{Experiment}
\label{sec:experiment}

\subsection{Experimental Setup}
\label{sec:setup}

\noindent\textbf{Benchmark.} 
We evaluate EvoSCM across three scientific domains.
(1) \textit{Physics}:
DiscoverPhysics~\citep{wiemann2026discoverphysics} requires agents to uncover the hidden dynamics of 22 noncanonical physical worlds through interactive experimentation.
(2) \textit{Chemistry \& Materials}:
LLEMA~\citep{abhyankar2026llema} tasks agents with multi-objective materials discovery across 14 material design challenges.
(3) \textit{Biology}:
ActiveSciBench-GRN~\citep{kabra2026llm} requires agents to recover the causal graph structure of 45 gene regulatory networks from interventional gene expression data through active experimentation.
Each task instance is evaluated over 5 independent random seeds.

\noindent\textbf{Base Agents and Evolution Methods.}
Each benchmark domain provides a baseline agent implementing the domain-specific experimental loop (hypothesis generation, experiment execution, and result interpretation)~\citep{wiemann2026discoverphysics,abhyankar2026llema,kabra2026llm}.
We equip each base agent with five evolution methods: GEPA~\citep{agrawal2026gepa} (reflective prompt evolution), ReasoningBank~\citep{ouyang2026reasoningbank} (reasoning memory distillation), ModelSMC~\citep{wahl2026probabilistic} (probabilistic model discovery), PiEvo~\citep{pu2026principle} (principle evolution via uncertainty minimization), and EvoSCM.
All methods share the same base agent, LLM backbone, and experimental budget per domain, differing only in how they evolve the agent's state from evidence.

\subsection{Evaluation}

\begin{table*}[t]
\centering
\caption{\textbf{Physical law discovery on DiscoverPhysics}~\citep{wiemann2026discoverphysics}. EvoSCM consistently outperforms all baselines in explanation score, prediction error, success rate, and experimental efficiency on both GPT-5.4~\citep{openai_gpt54} and GPT-5.5~\citep{openai_gpt55}.
}
\label{tab:physics}
\renewcommand{\arraystretch}{1.2}
\setlength{\tabcolsep}{6pt}
\resizebox{\textwidth}{!}{%
\begin{tabular}{llcccccccc}
\toprule
\textbf{Backbone} 
& \textbf{Method} 
& \textbf{$\langle$Explanation$\rangle$ $\uparrow$} 
& \textbf{norm $\langle$MSE$\rangle \downarrow$} 
& \textbf{pass@1 $\uparrow$} 
& \textbf{pass@2 $\uparrow$} 
& \textbf{pass@3 $\uparrow$} 
& \textbf{pass@4 $\uparrow$} 
& \textbf{pass@5 $\uparrow$} 
& \textbf{Episodes $\downarrow$} \\
\midrule

\multirow{6}{*}{GPT-5.4}
& Baseline
& 39.80\%
& 5.25e-1
& 1.86\%
& 3.88\%
& 5.60\%
& 7.14\%
& 9.09\%
& 1,045 \\

& GEPA
& 35.64\%
& 5.83e-1
& 0.00\%
& 0.00\%
& 0.00\%
& 0.00\%
& 0.00\%
& 976 \\

& ReasoningBank
& 36.55\%
& 1.84e0
& 5.45\%
& 10.00\%
& 13.64\%
& 16.36\%
& 18.18\%
& 908 \\

& ModelSMC
& 28.55\%
& 1.28e0
& 1.82\%
& 3.64\%
& 5.45\%
& 7.27\%
& 9.09\%
& 901 \\

& PiEvo
& 28.18\%
& 2.25e0
& 1.82\%
& 3.64\%
& 5.45\%
& 7.27\%
& 9.09\%
& 1,033 \\

\rowcolor{myblue}
\cellcolor{white}
& \textbf{EvoSCM (Ours)}
& \textbf{65.30\%}
& \textbf{2.34e-3}
& \textbf{38.08\%}
& \textbf{50.16\%}
& \textbf{53.74\%}
& \textbf{54.55\%}
& \textbf{54.55\%}
& \textbf{736} \\

\midrule

\multirow{6}{*}{GPT-5.5}
& Baseline
& 51.64\%
& 2.83e-2
& 7.27\%
& 13.64\%
& 19.09\%
& 23.64\%
& 27.27\%
& 2,206 \\

& GEPA
& 45.09\%
& 3.55e-2
& 9.09\%
& 14.55\%
& 17.27\%
& 18.18\%
& 18.18\%
& 2,062 \\

& ReasoningBank
& 51.27\%
& 2.78e-2
& 10.91\%
& 19.09\%
& 24.55\%
& 27.27\%
& 27.27\%
& 1,797 \\

& ModelSMC
& 43.27\%
& 1.83e-2
& 7.27\%
& 13.64\%
& 19.09\%
& 23.64\%
& 27.27\%
& 1,884 \\

& PiEvo
& 48.18\%
& 5.01e-2
& 9.09\%
& 16.36\%
& 21.82\%
& 25.45\%
& 27.27\%
& 1,845 \\

\rowcolor{myblue}
\cellcolor{white}
& \textbf{EvoSCM (Ours)}
& \textbf{75.09\%}
& \textbf{2.77e-4}
& \textbf{56.36\%}
& \textbf{62.73\%}
& \textbf{63.64\%}
& \textbf{63.64\%}
& \textbf{63.64\%}
& \textbf{1,685} \\

\bottomrule
\end{tabular}%
}
\end{table*}

\noindent\textbf{Physical Law Discovery.}
Table~\ref{tab:physics} reports results on DiscoverPhysics~\citep{wiemann2026discoverphysics}. Following its evaluation protocol, we measure mechanism explanation score (Explanation~$\uparrow$), normalized prediction error (norm~MSE~$\downarrow$), and per-world success rate (pass@$k$~$\uparrow$). We additionally report experimental efficiency (Episodes~$\downarrow$) to measure how quickly each method converges.
EvoSCM substantially outperforms all baselines on GPT-5.4~\citep{openai_gpt54} and GPT-5.5~\citep{openai_gpt55}, reducing prediction error by over two orders of magnitude while requiring fewer experimental episodes.
These results suggest that refining how an agent reasons is insufficient when its underlying scientific beliefs remain implicit, and that EvoSCM's gains stem from maintaining and revising an explicit causal model that grounds every prediction in a falsifiable structural commitment.

\begin{table*}[t]
\centering
\caption{\textbf{Chemistry \& Materials discovery on LLEMA}~\citep{abhyankar2026llema}. EvoSCM consistently outperforms all baselines in hit rate (H.R.,~\%) and stability (Stab.,~\%) based on Qwen3.6-35B-A3B~\citep{qwen36_35b_a3b}.
}
\label{tab:chemistry}
\setlength{\tabcolsep}{4pt}
\renewcommand{\arraystretch}{1.2}
\resizebox{\textwidth}{!}{%
\begin{tabular}{l*{14}{c}}
\toprule

\multirow{2}{*}{\textbf{Method}}
& \multicolumn{2}{c}{\makecell{\textbf{Wide-Bandgap}\\\textbf{Semicond.}}}
& \multicolumn{2}{c}{\makecell{\textbf{SAW/BAW}\\\textbf{Acoustic Substrates}}}
& \multicolumn{2}{c}{\makecell{\textbf{High-$k$}\\\textbf{Dielectrics}}}
& \multicolumn{2}{c}{\makecell{\textbf{Solid-State}\\\textbf{Electrolytes}}}
& \multicolumn{2}{c}{\makecell{\textbf{Piezo Energy}\\\textbf{Harvesters}}}
& \multicolumn{2}{c}{\makecell{\textbf{Transparent}\\\textbf{Conductors}}}
& \multicolumn{2}{c}{\makecell{\textbf{Insulating}\\\textbf{Dielectrics}}} \\
\cmidrule(lr){2-3}
\cmidrule(lr){4-5}
\cmidrule(lr){6-7}
\cmidrule(lr){8-9}
\cmidrule(lr){10-11}
\cmidrule(lr){12-13}
\cmidrule(lr){14-15}
& H.R $\uparrow$ & Stab. $\uparrow$
& H.R $\uparrow$ & Stab. $\uparrow$
& H.R $\uparrow$ & Stab. $\uparrow$
& H.R $\uparrow$ & Stab. $\uparrow$
& H.R $\uparrow$ & Stab. $\uparrow$
& H.R $\uparrow$ & Stab. $\uparrow$
& H.R $\uparrow$ & Stab. $\uparrow$ \\
\midrule

Baseline
& 3.33 & 0.83
& 45.00 & 0.00
& 0.00 & 0.00
& 5.00 & 1.67
& 32.50 & 0.00
& 0.00 & 0.00
& 0.83 & 0.00 \\

GEPA
& 10.00 & 0.00
& 38.33 & 0.00
& 1.67 & 0.00
& 2.50 & 0.00
& 35.83 & 0.00
& 0.00 & 0.00
& 2.50 & 0.00 \\

ReasoningBank
& 14.17 & 6.67
& 32.50 & 0.00
& 0.00 & 0.00
& 13.33 & 6.67
& 26.67 & 0.00
& 0.00 & 0.00
& 3.33 & 0.00 \\

ModelSMC
& 10.83 & 5.83
& 45.83 & 3.33
& 0.00 & 0.00
& 30.00 & 0.83
& 39.17 & 0.00
& 0.00 & 0.00
& 1.67 & 0.00 \\

PiEvo
& 20.83 & 5.83
& 40.83 & 0.00
& 1.67 & 0.00
& 22.50 & 15.00
& 54.17 & 0.00
& 0.00 & 0.00
& 0.83 & 0.00 \\

\midrule
\rowcolor{myblue}
\textbf{EvoSCM (Ours)}
& \textbf{21.67} & \textbf{15.00}
& \textbf{75.83} & \textbf{15.00}
& \textbf{5.00} & \textbf{5.00}
& \textbf{40.83} & \textbf{30.00}
& \textbf{69.17} & \textbf{6.67}
& \textbf{10.83} & \textbf{10.83}
& \textbf{6.67} & \textbf{5.00} \\

\midrule

\multirow{2}{*}{\textbf{Method}}
& \multicolumn{2}{c}{\makecell{\textbf{Photovoltaics}\\\textbf{Absorbers}}}
& \multicolumn{2}{c}{\makecell{\textbf{Hard Coating}\\\textbf{Materials}}}
& \multicolumn{2}{c}{\makecell{\textbf{Hard, Stiff}\\\textbf{Ceramics}}}
& \multicolumn{2}{c}{\makecell{\textbf{Aerospace}\\\textbf{Materials}}}
& \multicolumn{2}{c}{\makecell{\textbf{Acousto-optic}\\\textbf{Hybrids}}}
& \multicolumn{2}{c}{\makecell{\textbf{Low Density}\\\textbf{Structures}}}
& \multicolumn{2}{c}{\makecell{\textbf{Perovskite}\\\textbf{Oxides}}} \\
\cmidrule(lr){2-3}
\cmidrule(lr){4-5}
\cmidrule(lr){6-7}
\cmidrule(lr){8-9}
\cmidrule(lr){10-11}
\cmidrule(lr){12-13}
\cmidrule(lr){14-15}
& H.R $\uparrow$ & Stab. $\uparrow$
& H.R $\uparrow$ & Stab. $\uparrow$
& H.R $\uparrow$ & Stab. $\uparrow$
& H.R $\uparrow$ & Stab. $\uparrow$
& H.R $\uparrow$ & Stab. $\uparrow$
& H.R $\uparrow$ & Stab. $\uparrow$
& H.R $\uparrow$ & Stab. $\uparrow$ \\
\midrule

Baseline
& 0.00 & 0.00
& 0.00 & 0.00
& 44.17 & 0.00
& 40.00 & 0.00
& 7.50 & 0.00
& 0.00 & 0.00
& 1.67 & 0.00 \\

GEPA
& 0.00 & 0.00
& 2.50 & 0.00
& 41.67 & 0.00
& 37.50 & 0.00
& 8.33 & 0.00
& 0.00 & 0.00
& 0.00 & 0.00 \\

ReasoningBank
& 0.00 & 0.00
& 0.00 & 0.00
& 50.00 & 0.00
& 42.50 & 0.00
& 12.50 & 0.00
& 0.00 & 0.00
& 0.83 & 0.00 \\

ModelSMC
& 0.00 & 0.00
& 0.00 & 0.00
& 25.00 & 0.00
& 28.33 & 0.00
& 12.50 & 0.00
& 0.00 & 0.00
& 0.00 & 0.00 \\

PiEvo
& 1.67 & 0.00
& 0.00 & 0.00
& 61.67 & 0.00
& 47.50 & 0.00
& 9.17 & 0.00
& 1.67 & 1.67
& 0.00 & 0.00 \\

\midrule
\rowcolor{myblue}
\textbf{EvoSCM (Ours)}
& \textbf{11.67} & \textbf{7.50}
& \textbf{24.17} & \textbf{23.33}
& \textbf{68.33} & \textbf{8.33}
& \textbf{60.00} & \textbf{8.33}
& \textbf{20.83} & \textbf{6.67}
& \textbf{5.00} & \textbf{5.00}
& \textbf{7.50} & \textbf{4.17} \\

\bottomrule
\end{tabular}%
}
\end{table*}

\noindent\textbf{Chemistry \& Materials Discovery.}
Table~\ref{tab:chemistry} reports results on LLEMA~\citep{abhyankar2026llema} across 14 material design tasks based on Qwen3.6-35B-A3B~\citep{qwen36_35b_a3b}. Following its evaluation protocol, we measure hit rate (H.R.~$\uparrow$), the percentage of discovered materials satisfying target property constraints, and stability (Stab.~$\uparrow$), the percentage of discovered materials remaining valid under perturbation.
EvoSCM achieves the highest hit rate and stability on all 14 tasks.
Most baselines achieve zero stability on the majority of tasks, indicating that their discovered materials are brittle and fail under perturbation, while EvoSCM achieves nonzero stability on every task.
This suggests that an explicit causal model of material property relationships enables EvoSCM to discover candidates grounded in structural mechanisms rather than surface-level heuristics.

\begin{table*}[t]
\centering
\caption{\textbf{Biological network inference on ActiveSciBench-GRN}~\citep{kabra2026llm}. EvoSCM consistently outperforms baselines in edge F1 (\%), exact graph accuracy (\%) and sign accuracy (\%) on Qwen3.6-35B-A3B~\citep{qwen36_35b_a3b} and GPT-5.6-Luna~\citep{openai_gpt56}.
}
\label{tab:biology}
\setlength{\tabcolsep}{6pt}
\renewcommand{\arraystretch}{1.2}
\resizebox{\textwidth}{!}{%
\begin{tabular}{lcccccc}
    \toprule
    \multirow{2}{*}{\textbf{Method}}
    & \multicolumn{3}{c}{Qwen3.6-35B-A3B}
    & \multicolumn{3}{c}{GPT-5.6-Luna} \\
    \cmidrule(lr){2-4}
    \cmidrule(lr){5-7}
    & \textbf{Edge F1 $\uparrow$}
    & \textbf{Exact Graph Accuracy $\uparrow$}
    & \textbf{Sign Accuracy $\uparrow$}
    & \textbf{Edge F1 $\uparrow$}
    & \textbf{Exact Graph Accuracy $\uparrow$}
    & \textbf{Sign Accuracy $\uparrow$} \\
    \midrule

    Baseline
    & 81.71 & 28.15 & 95.44
    & 81.49 & 18.52 & 98.32 \\

    GEPA
    & 84.28 & 36.30 & 98.09
    & 85.48 & 25.19 & 98.30 \\

    ReasoningBank
    & 84.10 & 35.56 & 97.05
    & 86.04 & 34.07 & 99.15 \\

    ModelSMC
    & 76.57 & 28.89 & 96.74
    & 82.59 & 21.48 & 97.29 \\

    PiEvo
    & 77.79 & 25.93 & 97.58
    & 77.38 & 17.41 & 97.79 \\

    \midrule

    \rowcolor{myblue}
    \textbf{EvoSCM (Ours)}
    & \textbf{96.12}
    & \textbf{70.37}
    & \textbf{99.85}
    & \textbf{96.94}
    & \textbf{77.04}
    & \textbf{99.82} \\

    \bottomrule
\end{tabular}
}
\end{table*}

\noindent\textbf{Biological Network Inference.}
Table~\ref{tab:biology} reports results on ActiveSciBench-GRN~\citep{kabra2026llm}, where agents must recover a signed, directed causal graph of gene regulatory networks from budget-limited perturbation experiments.
Following its evaluation protocol, we measure edge F1~($\uparrow$), exact graph accuracy~($\uparrow$), and sign accuracy~($\uparrow$), evaluating the agent's ability to recover correct regulatory edges, complete network topologies, and edge polarities, respectively.
EvoSCM substantially outperforms all baselines on every metric, on both Qwen3.6-35B-A3B~\citep{qwen36_35b_a3b} and GPT-5.6-Luna~\citep{openai_gpt56}.
The most notable gap is in exact graph accuracy, indicating that EvoSCM recovers complete network topologies far more reliably than baselines that improve individual edge predictions without maintaining a coherent graph-level hypothesis.

\subsection{Analysis}

\begin{table*}[t]
\centering
\caption{\textbf{Cross-model transferability results on DiscoverPhysics}~\citep{wiemann2026discoverphysics}. SCMs evolved by GPT transfer seamlessly to Qwen3.6-35B-A3B, substantially improving its performance.
}
\label{tab:transfer_results}
\renewcommand{\arraystretch}{1.2}
\setlength{\tabcolsep}{6pt}
\resizebox{\textwidth}{!}{%
\begin{tabular}{lccccccc}
\toprule
\textbf{Method} & \textbf{$\langle$Explanation$\rangle \uparrow$} & \textbf{norm $\langle$MSE$\rangle \downarrow$} & \textbf{pass@1 $\uparrow$} & \textbf{pass@2 $\uparrow$} & \textbf{pass@3 $\uparrow$} & \textbf{pass@4 $\uparrow$} & \textbf{pass@5 $\uparrow$} \\
\midrule
Qwen3.6 Baseline & 30.73\% & 1.90e-1 & 0.00\% & 0.00\% & 0.00\% & 0.00\% & 0.00\% \\
Qwen3.6 + Qwen3.6 SCM & 48.91\% & 6.31e-2 & 14.95\% & 26.07\% & 32.92\% & 36.36\% & 36.36\% \\
Qwen3.6 + GPT-5.4 SCM & 57.64\% & 2.85e-2 & 16.40\% & 23.43\% & 26.41\% & 27.27\% & 27.27\% \\
Qwen3.6 + GPT-5.5 SCM & 62.55\% & 8.69e-3 & 34.87\% & 39.52\% & 41.96\% & 43.72\% & 45.45\% \\
Qwen3.6 + GPT-5.6-Sol SCM & 74.00\% & 1.06e-4 & 39.85\% & 49.76\% & 57.14\% & 61.87\% & 63.64\% \\
\bottomrule
\end{tabular}%
}
\end{table*}

\noindent\textbf{Cross-Model SCM Transferability.}
A key advantage of representing scientific knowledge explicitly as an SCM is its portability across model architectures, as shown in Table~\ref{tab:transfer_results}.
We use Qwen3.6-35B-A3B~\citep{qwen36_35b_a3b}, which fails outright on DiscoverPhysics, as the transfer target. Equipping Qwen3.6 with an SCM it evolved itself already recovers substantial capability, and injecting SCMs evolved by stronger models (GPT-5.4~\citep{openai_gpt54}, GPT-5.5~\citep{openai_gpt55}, and GPT-5.6-Sol~\citep{openai_gpt56}) pushes explanation accuracy and prediction error further in the same direction, with gains scaling monotonically with the strength of the source model.
This confirms that the evolved SCMs capture reusable scientific knowledge beyond any particular model's reasoning process and can be transferred directly across base models.

\begin{figure}[t]
    \centering
    \includegraphics[width=\linewidth]{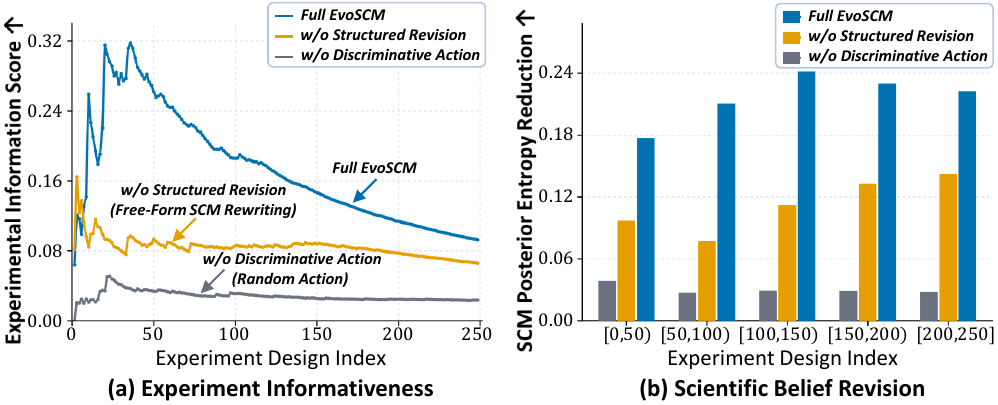}
    \vspace{-10pt}
    \caption{\textbf{Experimentation and belief revision analysis on DiscoverPhysics}~\citep{wiemann2026discoverphysics}.
    (a)~\emph{Experiment Informativeness}: EvoSCM designs experiments with higher information scores~\citep{lindley1956measure,mackay1992information}. The score peaks early when the hypothesis population is most diverse and declines as competing hypotheses converge, leaving less disagreement to exploit.
    (b)~\emph{Scientific Belief Revision}: EvoSCM achieves consistently larger posterior entropy reductions~\citep{chaloner1995bayesian,mackay1992information}, confirming more effective belief revision.
    }
    \label{fig:analysis1}
\end{figure}

\noindent\textbf{Experimentation and Belief Revision Analysis.}
Figure~\ref{fig:analysis1} examines the two directions of EvoSCM's discovery loop, comparing EvoSCM against two ablations: w/o discriminative action (random action) and w/o structured revision (free-form SCM rewriting).
EvoSCM designs experiments with substantially higher information scores~\citep{lindley1956measure,mackay1992information}, demonstrating that the SCM population guides the agent toward interventions that maximally discriminate between competing hypotheses.
These informative experiments translate into consistently larger SCM posterior entropy reductions~\citep{chaloner1995bayesian,mackay1992information}: the agent's uncertainty over the correct causal model decreases faster per experiment, concentrating posterior weight on hypotheses that best explain the evidence and driving more effective scientific belief revision.

\begin{figure}[t]
    \centering
    \includegraphics[width=\linewidth]{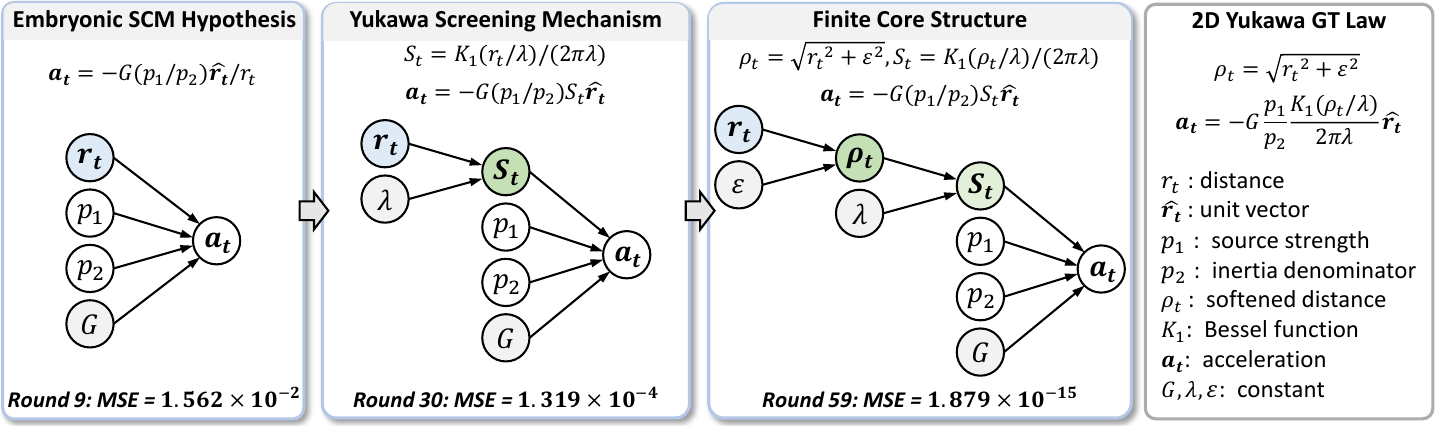}
    \caption{\textbf{SCM hypothesis evolution on the Yukawa world in
  DiscoverPhysics}~\citep{wiemann2026discoverphysics}.
  EvoSCM progressively recovers the 2D Yukawa ground-truth law
  through physically meaningful intermediate states, rather than
  merely fitting parameters along a smooth refinement path.
  (1)~\emph{Round~9}: the agent starts from an embryonic Coulomb-like 
  hypothesis ($\boldsymbol{a}_t \propto 1/r_t$);
  (2)~\emph{Round~30}: a revision introduces a screening factor
  $S_t = K_1(r_t/\lambda)/(2\pi\lambda)$, coinciding with the
  Yukawa screening mechanism~\citep{yukawa1935interaction},
  resolving intermediate-range overprediction;
  (3)~\emph{Round~59}: a further revision introduces a softened
  distance $\rho_t = \sqrt{r_t^2 + \varepsilon^2}$, coinciding
  with the finite core model~\citep{jastrow1951nucleon},
  resolving residual short-range discrepancies.
  }
    \label{fig:analysis2}
\end{figure}

\noindent\textbf{SCM Hypothesis Evolution Analysis.}
Figure~\ref{fig:analysis2} traces the evolution of EvoSCM's SCM
hypothesis on the Yukawa world, from an embryonic Coulomb-like hypothesis
at Round~9 ($\boldsymbol{a}_t \propto 1/r_t$) to the ground-truth Yukawa law.
The intermediate states along this trajectory are not
arbitrary but physically meaningful, each resolving a distinct
class of prediction failures.
At Round~30, the agent introduces a screening factor
$S_t = K_1(r_t/\lambda)/(2\pi\lambda)$, a structure coinciding
with the Yukawa screening mechanism~\citep{yukawa1935interaction},
which resolves the systematic overprediction at intermediate
distances.
At Round~59, a further revision introduces a softened distance
$\rho_t = \sqrt{r_t^2 + \varepsilon^2}$, coinciding with the
finite core model~\citep{jastrow1951nucleon}, which resolves the
residual short-range discrepancies that screening alone cannot
explain.
This observation suggests that the evidence-driven causal model
evolution does not merely fit parameters along a smooth loss
landscape but discovers structurally distinct intermediate
hypotheses, each capturing a qualitatively new aspect of the
underlying law.

\begin{table*}[t]
\centering
\renewcommand{\arraystretch}{1.2}

\begin{minipage}[t]{0.61\textwidth}
\centering
\caption{\textbf{Ablation study on EvoSCM.} 
We evaluate SCM representation, discriminative action, and structured revision.
}
\label{tab:ablation1}
\renewcommand{\arraystretch}{1.2}
\setlength{\tabcolsep}{1pt}
\resizebox{\linewidth}{!}{%
\begin{tabular}{lcc}
\toprule
\textbf{Method}
& \textbf{$\langle$Explanation$\rangle$ $\uparrow$}
& \textbf{norm $\langle$MSE$\rangle$ $\downarrow$} \\
\midrule

Full EvoSCM
& \textbf{75.09\%}
& \textbf{2.77e-4} \\

w/o SCM (Free-Form Textual Hypotheses)
& 48.18\%
& 7.82e-2\\

w/o Discriminative Action (Random Action)
& 47.45\%
& 1.05e-1 \\

w/o Structured Revision (Free-Form SCM Rewriting)
& 70.73\%
& 5.45e-4 \\

\bottomrule
\end{tabular}%
}
\end{minipage}
\hfill
\begin{minipage}[t]{0.38\textwidth}
\centering
\caption{\textbf{Ablation study on the SCM hypothesis population size} ($K$).
}
\label{tab:ablation2}
\renewcommand{\arraystretch}{1.2}
\setlength{\tabcolsep}{4pt}
\resizebox{\linewidth}{!}{%
\begin{tabular}{ccc}
\toprule
\textbf{Population Size}
& \textbf{$\langle$Explanation$\rangle$ $\uparrow$}
& \textbf{norm $\langle$MSE$\rangle$ $\downarrow$} \\
\midrule

$K=1$
& 63.27\%
& 6.78e-3 \\

$K=4$
& 74.86\%
& 2.59e-4 \\

$K=6$
& \textbf{75.09\%}
& \textbf{2.77e-4} \\

$K=8$
& 75.01\%
& 2.71e-4 \\

\bottomrule
\end{tabular}%
}
\end{minipage}
\end{table*}

\noindent\textbf{Ablation Study on EvoSCM.}
Table~\ref{tab:ablation1} evaluates the contribution of each core component. Replacing SCMs with free-form textual hypotheses causes the largest drop, highlighting the value of an explicit causal representation. Removing discriminative action selection also substantially degrades performance, while free-form SCM rewriting yields a smaller but consistent decline, supporting targeted hypothesis revision. Table~\ref{tab:ablation2} studies population size: performance improves from $K\!=\!1$ to $K\!=\!4$ and remains stable for larger $K$, suggesting that maintaining several competing hypotheses is beneficial while EvoSCM is not sensitive to larger populations.

\section{Conclusion}

We present {EvoSCM}, a framework that represents scientific beliefs as explicit and persistent structural causal models, enabling scientific agents to revise not only how they reason, but also what they believe about the world. EvoSCM maintains a population of competing SCM hypotheses and evolves them through a closed loop of abduction, intervention, induction, and deduction, supporting a structured, falsifiable, and cumulatively revisable process of scientific discovery. Across physics, chemistry, and biology, EvoSCM consistently improves scientific discovery over baseline agents and existing evolution methods, yielding more accurate explanations and predictions while using experimental budgets more effectively. The evolved SCMs transfer across base architectures, suggesting that the resulting causal models encode reusable scientific knowledge beyond the reasoning process of any particular model. Together, these results support explicit causal model evolution as a promising direction for scientific agents that refine their understanding through experimentation.

\textit{Limitation and Future Direction.}
Our evaluation focuses on simulated scientific environments. Extending EvoSCM to real-world settings, where observations are noisy, interventions are costly, and ground truth is unavailable, remains an important next step.

\section*{AI use statement}

Large language models are an integral component of the proposed EvoSCM framework and its experimental evaluation. EvoSCM uses LLMs as the reasoning backbone of scientific agents, performing abductive inference, causal experiment design, inductive rule extraction, structural revision, and deductive validation, as described in Sections~\ref{sec:experiment_design} and~\ref{sec:revision}. We evaluate GPT-5.4, GPT-5.5, GPT-5.6-Luna, GPT-5.6-Sol, and Qwen3.6-35B-A3B as LLM backbones, executing the full scientific discovery loop on three benchmarks (DiscoverPhysics, LLEMA, and ActiveSciBench-GRN), as described in Section~\ref{sec:experiment}. All LLM-generated scientific outputs, including causal graphs, structural equations, predictions, experimental decisions, and revision decisions, are quantitatively evaluated against the corresponding ground-truth benchmarks. These uses of LLMs are intrinsic to the EvoSCM system studied and evaluated in this work.

Separately, generative AI tools were used to assist with polishing portions of the manuscript for clarity and presentation. Outside the execution and evaluation of the proposed EvoSCM framework, generative AI tools were not used to determine the core methodological contributions or to generate or alter the reported experimental results. All AI-assisted text was reviewed, revised where appropriate, and verified by the authors for accuracy, correctness, and originality. The authors take full responsibility for the final content of this work, including all text, claims, methods, experimental results, analyses, and artifacts produced with the aid of generative AI.

\bibliography{iclr2027_conference}
\bibliographystyle{iclr2027_conference}

\clearpage
\appendix
\section*{Appendix}

\startcontents[appendix]

\vspace{1em}
{\large\scshape Table of Contents}
\vspace{0.8em}

\printcontents[appendix]{}{1}{%
    \setcounter{tocdepth}{2}%
}

\vspace{1.5em}

\clearpage

\section{Additional Results}

\subsection{Additional Results on DiscoverPhysics}

Table~\ref{tab:physics_gpt56sol} reports additional DiscoverPhysics~\citep{wiemann2026discoverphysics} results using GPT-5.6-Sol~\citep{openai_gpt56} as the base model.
EvoSCM outperforms the baseline across all metrics, improving explanation score from 63.64\% to 79.13\%, reducing prediction error by over five times, and requiring roughly one-third fewer experimental episodes (3{,}543 vs.\ 5{,}296).
EvoSCM nevertheless produces consistent gains on top of this stronger base, confirming that its improvements are not limited to weaker models and that explicit causal model evolution provides complementary benefits beyond what a more capable LLM backbone alone can achieve.

\begin{table*}[h]
\centering
\caption{\textbf{Physical law discovery on DiscoverPhysics}~\citep{wiemann2026discoverphysics}. EvoSCM consistently outperforms all baselines in explanation score, prediction error, success rate, and experimental efficiency on GPT-5.6-Sol~\citep{openai_gpt56}.
}
\label{tab:physics_gpt56sol}
\renewcommand{\arraystretch}{1.2}
\setlength{\tabcolsep}{6pt}
\resizebox{\textwidth}{!}{%
\begin{tabular}{lcccccccc}
\toprule
\textbf{Method} 
& \textbf{$\langle$Explanation$\rangle$ $\uparrow$} 
& \textbf{norm $\langle$MSE$\rangle \downarrow$} 
& \textbf{pass@1 $\uparrow$} 
& \textbf{pass@2 $\uparrow$} 
& \textbf{pass@3 $\uparrow$} 
& \textbf{pass@4 $\uparrow$} 
& \textbf{pass@5 $\uparrow$} 
& \textbf{Episodes $\downarrow$} \\
\midrule

Baseline
& 63.64\%
& 7.79e-4
& 34.76\%
& 49.37\%
& 57.51\%
& 61.68\%
& 63.64\%
& 5296 \\

\rowcolor{myblue}
\textbf{EvoSCM (Ours)}
& \textbf{79.13\%}
& \textbf{9.96e-5}
& \textbf{43.82\%}
& \textbf{61.39\%}
& \textbf{68.30\%}
& \textbf{70.96\%}
& \textbf{72.73\%}
& \textbf{3543} \\

\bottomrule
\end{tabular}%
}
\end{table*}

\subsection{Additional Results on LLEMA}

Table~\ref{tab:chemistry_gpt56luna} reports additional LLEMA~\citep{abhyankar2026llema} results using GPT-5.6-Luna~\citep{openai_gpt56} as the base model, comparing EvoSCM against the LLEMA baseline agent.
EvoSCM achieves a higher hit rate and stability on all 14 material design tasks, consistent with the Qwen3.6-35B-A3B results reported in the main paper (Table~\ref{tab:chemistry}).
The stability advantage is again particularly notable: the baseline achieves zero stability on 10 of 14 tasks, while EvoSCM achieves nonzero stability on every task, with the largest margins on SAW/BAW Acoustic Substrates (19.17\% vs.\ 0.00\%) and Solid-State Electrolytes (29.17\% vs.\ 1.67\%).
These results confirm that EvoSCM's improvements generalize across LLM backbones and are not specific to a particular base architecture. 

\begin{table*}[h]
\centering
\caption{\textbf{Chemistry \& Materials discovery on LLEMA}~\citep{abhyankar2026llema}. EvoSCM consistently outperforms the LLEMA Baseline in hit rate (H.R.,~\%) and stability (Stab.,~\%) based on GPT-5.6-Luna~\citep{openai_gpt56}.
}
\label{tab:chemistry_gpt56luna}
\setlength{\tabcolsep}{4pt}
\renewcommand{\arraystretch}{1.3}
\resizebox{\textwidth}{!}{%
\begin{tabular}{l*{14}{c}}
\toprule

\multirow{2}{*}{\textbf{Method}}
& \multicolumn{2}{c}{\makecell{\textbf{Wide-Bandgap}\\\textbf{Semicond.}}}
& \multicolumn{2}{c}{\makecell{\textbf{SAW/BAW}\\\textbf{Acoustic Substrates}}}
& \multicolumn{2}{c}{\makecell{\textbf{High-$k$}\\\textbf{Dielectrics}}}
& \multicolumn{2}{c}{\makecell{\textbf{Solid-State}\\\textbf{Electrolytes}}}
& \multicolumn{2}{c}{\makecell{\textbf{Piezo Energy}\\\textbf{Harvesters}}}
& \multicolumn{2}{c}{\makecell{\textbf{Transparent}\\\textbf{Conductors}}}
& \multicolumn{2}{c}{\makecell{\textbf{Insulating}\\\textbf{Dielectrics}}} \\
\cmidrule(lr){2-3}
\cmidrule(lr){4-5}
\cmidrule(lr){6-7}
\cmidrule(lr){8-9}
\cmidrule(lr){10-11}
\cmidrule(lr){12-13}
\cmidrule(lr){14-15}
& H.R $\uparrow$ & Stab. $\uparrow$
& H.R $\uparrow$ & Stab. $\uparrow$
& H.R $\uparrow$ & Stab. $\uparrow$
& H.R $\uparrow$ & Stab. $\uparrow$
& H.R $\uparrow$ & Stab. $\uparrow$
& H.R $\uparrow$ & Stab. $\uparrow$
& H.R $\uparrow$ & Stab. $\uparrow$ \\
\midrule

Baseline
& 10.00 & 3.33
& 29.17 & 0.00
& 0.00 & 0.00
& 6.67 & 1.67
& 54.17 & 0.00
& 1.67 & 1.67
& 0.83 & 0.00 \\

\midrule
\rowcolor{myblue}
\textbf{EvoSCM (Ours)}
& \textbf{16.67} & \textbf{10.00}
& \textbf{64.17} & \textbf{19.17}
& \textbf{8.33} & \textbf{6.67}
& \textbf{31.67} & \textbf{29.17}
& \textbf{79.17} & \textbf{5.00}
& \textbf{7.50} & \textbf{7.50}
& \textbf{6.67} & \textbf{4.17} \\

\midrule

\multirow{2}{*}{\textbf{Method}}
& \multicolumn{2}{c}{\makecell{\textbf{Photovoltaics}\\\textbf{Absorbers}}}
& \multicolumn{2}{c}{\makecell{\textbf{Hard Coating}\\\textbf{Materials}}}
& \multicolumn{2}{c}{\makecell{\textbf{Hard, Stiff}\\\textbf{Ceramics}}}
& \multicolumn{2}{c}{\makecell{\textbf{Aerospace}\\\textbf{Materials}}}
& \multicolumn{2}{c}{\makecell{\textbf{Acousto-optic}\\\textbf{Hybrids}}}
& \multicolumn{2}{c}{\makecell{\textbf{Low Density}\\\textbf{Structures}}}
& \multicolumn{2}{c}{\makecell{\textbf{Perovskite}\\\textbf{Oxides}}} \\
\cmidrule(lr){2-3}
\cmidrule(lr){4-5}
\cmidrule(lr){6-7}
\cmidrule(lr){8-9}
\cmidrule(lr){10-11}
\cmidrule(lr){12-13}
\cmidrule(lr){14-15}
& H.R $\uparrow$ & Stab. $\uparrow$
& H.R $\uparrow$ & Stab. $\uparrow$
& H.R $\uparrow$ & Stab. $\uparrow$
& H.R $\uparrow$ & Stab. $\uparrow$
& H.R $\uparrow$ & Stab. $\uparrow$
& H.R $\uparrow$ & Stab. $\uparrow$
& H.R $\uparrow$ & Stab. $\uparrow$ \\
\midrule

Baseline
& 0.00 & 0.00
& 0.83 & 0.00
& 25.83 & 0.00
& 47.50 & 0.00
& 8.33 & 0.83
& 0.83 & 0.83
& 1.67 & 0.00 \\

\midrule
\rowcolor{myblue}
\textbf{EvoSCM (Ours)}
& \textbf{8.33} & \textbf{7.50}
& \textbf{8.33} & \textbf{6.67}
& \textbf{58.33} & \textbf{9.17}
& \textbf{62.50} & \textbf{5.83}
& \textbf{17.50} & \textbf{6.67}
& \textbf{5.00} & \textbf{5.00}
& \textbf{7.50} & \textbf{4.17} \\

\bottomrule
\end{tabular}%
}
\end{table*}

\subsection{Additional Results on ActiveSciBench-GRN}

Tables~\ref{tab:grn1} and~\ref{tab:grn2} report per-regime results on ActiveSciBench-GRN~\citep{kabra2026llm} across three predefined kinetic regimes (Easy, Medium, Hard) on Qwen3.6-35B-A3B~\citep{qwen36_35b_a3b} and GPT-5.6-Luna~\citep{openai_gpt56}, respectively.
The regimes reflect progressively sharper and more nonlinear parameter settings: Easy tasks exhibit quasi-linear responses where small perturbations reveal the graph clearly, Medium tasks introduce saturation effects that obscure weak edges, and Hard tasks feature bistability and switching behavior where the graph is identifiable only through carefully designed multi-node perturbations~\citep{kabra2026llm}.
EvoSCM substantially outperforms all baselines on edge F1 and exact graph accuracy across every regime and both backbones, with particularly large margins in exact graph accuracy, the most demanding metric requiring every edge to be recovered correctly. On Qwen3.6, EvoSCM achieves perfect sign accuracy on Medium and Hard regimes and near-perfect on Easy, while on GPT-5.6-Luna it achieves competitive sign accuracy across all regimes, with ReasoningBank matching or slightly exceeding on two regimes but trailing substantially on edge F1 and exact graph accuracy. EvoSCM's performance remains stable across regimes: the gap between Easy and Hard is minimal on both backbones, whereas several baselines (particularly ModelSMC and PiEvo) degrade on harder regimes. This robustness suggests that EvoSCM's causal experiment design is particularly effective in the Hard regime, where the correct graph structure can only be identified through targeted multi-node interventions rather than simple single-variable perturbations, exactly the setting where population-level disagreement guides the agent toward the most revealing experiments.

\begin{table*}[h]
\centering
\caption{\textbf{Biological network inference across three predefined kinetic regimes on ActiveSciBench-GRN}~\citep{kabra2026llm}. EvoSCM consistently outperforms baselines in edge F1 (\%), exact graph accuracy (\%) and sign accuracy (\%) on Qwen3.6-35B-A3B~\citep{qwen36_35b_a3b}.
}
\label{tab:grn1}
\renewcommand{\arraystretch}{1.2}
\setlength{\tabcolsep}{3pt}
\resizebox{\textwidth}{!}{%
\begin{tabular}{lccccccccc}
\toprule
\multirow{2}{*}{\textbf{Method}}
& \multicolumn{3}{c}{\textbf{Easy}}
& \multicolumn{3}{c}{\textbf{Medium}}
& \multicolumn{3}{c}{\textbf{Hard}} \\
\cmidrule(lr){2-4}
\cmidrule(lr){5-7}
\cmidrule(lr){8-10}
& \textbf{Edge F1 $\uparrow$}
& \textbf{Exact Graph $\uparrow$}
& \textbf{Sign $\uparrow$}
& \textbf{Edge F1 $\uparrow$}
& \textbf{Exact Graph $\uparrow$}
& \textbf{Sign $\uparrow$}
& \textbf{Edge F1 $\uparrow$}
& \textbf{Exact Graph $\uparrow$}
& \textbf{Sign $\uparrow$} \\
\midrule

Baseline
& 81.59 & 33.33 & 95.04
& 82.24 & 33.33 & 95.93
& 81.29 & 17.78 & 95.37 \\

GEPA
& 83.63 & 31.11 & 96.67
& 85.00 & 44.44 & 99.44
& 84.20 & 33.33 & 98.15 \\

ReasoningBank
& 82.40 & 33.33 & 96.04
& 85.59 & 44.44 & 97.70
& 84.30 & 28.89 & 97.41 \\

ModelSMC
& 77.25 & 33.33 & 95.85
& 77.87 & 28.89 & 97.33
& 74.60 & 24.44 & 97.04 \\

PiEvo
& 77.73 & 26.67 & 99.00
& 77.10 & 24.44 & 97.70
& 78.52 & 26.67 & 96.04 \\

\rowcolor{myblue}
\textbf{EvoSCM (Ours)}
& \textbf{96.19} & \textbf{68.89} & \textbf{99.56}
& \textbf{96.43} & \textbf{71.11} & \textbf{100.00}
& \textbf{95.74} & \textbf{71.11} & \textbf{100.00} \\

\bottomrule
\end{tabular}%
}
\end{table*}

\begin{table*}[h]
\centering
\caption{\textbf{Biological network inference across three predefined kinetic regimes on ActiveSciBench-GRN}~\citep{kabra2026llm}. EvoSCM substantially improves edge F1 (\%) and exact graph accuracy (\%), achieving competitive sign accuracy (\%) on GPT-5.6-Luna~\citep{openai_gpt56}.
}
\label{tab:grn2}
\renewcommand{\arraystretch}{1.2}
\setlength{\tabcolsep}{3pt}
\resizebox{\textwidth}{!}{%
\begin{tabular}{lccccccccc}
\toprule
\multirow{2}{*}{\textbf{Method}}
& \multicolumn{3}{c}{\textbf{Easy}}
& \multicolumn{3}{c}{\textbf{Medium}}
& \multicolumn{3}{c}{\textbf{Hard}} \\
\cmidrule(lr){2-4}
\cmidrule(lr){5-7}
\cmidrule(lr){8-10}
& \textbf{Edge F1 $\uparrow$}
& \textbf{Exact Graph $\uparrow$}
& \textbf{Sign $\uparrow$}
& \textbf{Edge F1 $\uparrow$}
& \textbf{Exact Graph $\uparrow$}
& \textbf{Sign $\uparrow$}
& \textbf{Edge F1 $\uparrow$}
& \textbf{Exact Graph $\uparrow$}
& \textbf{Sign $\uparrow$} \\
\midrule

Baseline
& 80.02 & 13.33 & 97.52
& 81.28 & 24.44 & 99.00
& 83.17 & 17.78 & 98.44 \\

GEPA
& 85.62 & 20.00 & 97.56
& 85.17 & 26.67 & 98.89
& 85.65 & 28.89 & 98.44 \\

ReasoningBank
& 85.04 & 28.89 & 97.89
& 86.95 & 37.78 & \textbf{100.00}
& 86.13 & 35.56 & \textbf{99.56} \\

ModelSMC
& 82.75 & 17.78 & 98.26
& 85.09 & 28.89 & 97.11
& 79.93 & 17.78 & 96.52 \\

PiEvo
& 78.28 & 12.22 & 99.00
& 76.71 & 18.89 & 96.78
& 77.14 & 21.11 & 97.59 \\

\rowcolor{myblue}
\textbf{EvoSCM (Ours)}
& \textbf{96.81} & \textbf{80.00} & \textbf{99.96}
& \textbf{97.70} & \textbf{80.00} & 99.96
& \textbf{96.31} & \textbf{71.11} & \textbf{99.56} \\

\bottomrule
\end{tabular}%
}
\end{table*}

\subsection{LLM Token Usage}

\begin{figure}[h]
    \centering
    \includegraphics[width=\linewidth]{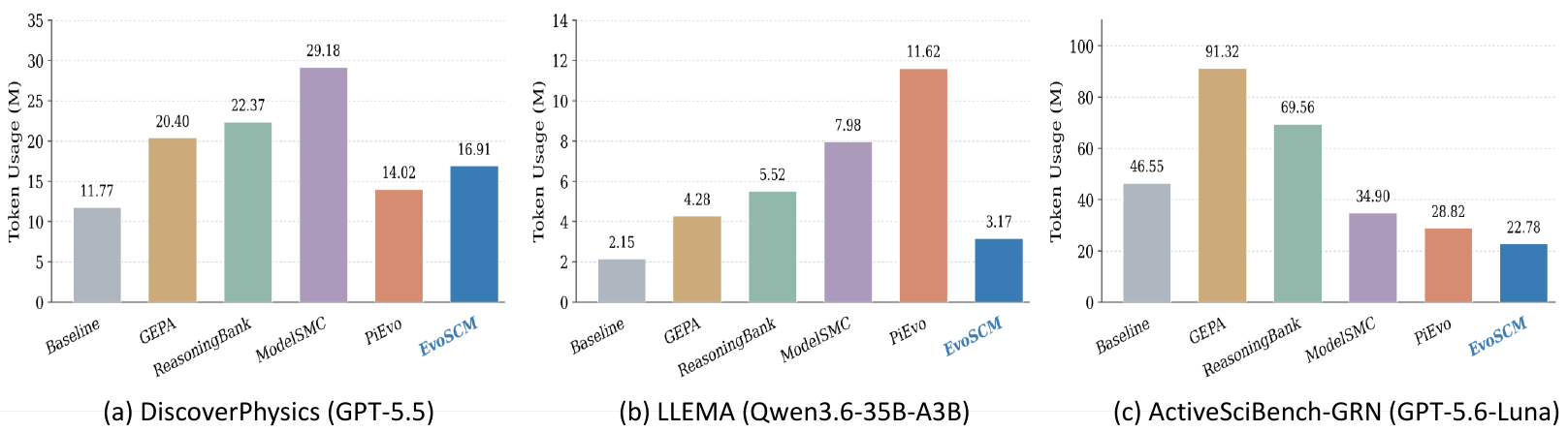}
    \caption{\textbf{LLM token usage (millions) across three benchmarks.} EvoSCM achieves the strongest scientific discovery performance (Tables~\ref{tab:physics},~\ref{tab:chemistry}, and~\ref{tab:biology}) while maintaining competitive token usage compared to other evolution methods across all three domains.}
    \label{fig:token}
\end{figure}

Figure~\ref{fig:token} reports the total LLM token consumption for each method across the three benchmarks.
Despite achieving consistently superior discovery performance, EvoSCM maintains moderate to low token usage relative to other evolution methods.
On LLEMA, EvoSCM consumes the fewest tokens among all evolution methods, using only marginally more than the unevolved baseline.
On ActiveSciBench-GRN, EvoSCM uses fewer tokens than even the baseline agent, likely because its more causal experiment design and targeted revisions lead to faster convergence, reducing the total
number of reasoning steps required.
On DiscoverPhysics, EvoSCM uses more tokens than the baseline and PiEvo but substantially fewer than GEPA, ReasoningBank, and ModelSMC.
This efficiency can be attributed to the structured SCM representation: causal graphs, structural equations, and parameters encode scientific knowledge compactly, whereas free-form textual hypotheses, reasoning memories, and verbose model descriptions require substantially more tokens to represent and manipulate.
Overall, these results indicate that EvoSCM's performance gains are not achieved by simply consuming more computational resources but rather by using tokens more effectively through structured causal representations and targeted revisions.

\section{Additional Analysis}

\subsection{Additional Ablation Study}

Table~\ref{tab:additional_ablation} extends the main ablation study
(Table~\ref{tab:ablation1}) with three additional ablations that
complete the coverage of EvoSCM's discovery loop stages.

\noindent\textbf{w/o Induction (No Correction Rules).}
Raw prediction-observation discrepancies are passed directly to
the Revision step without being distilled into correction rules.
The LLM must simultaneously identify the failure pattern and
determine the appropriate structural edit, rather than receiving a
distilled directive that separates diagnosis from repair.
Performance degrades moderately, confirming that the inductive
abstraction step produces more targeted revisions by decoupling
\emph{what went wrong} from \emph{how to fix it}.

\noindent\textbf{w/o Pre-committed Prediction (Only Observations).}
The agent observes experimental outcomes before formulating its
explanation, rather than committing to hypothesis-specific 
predictions in advance. Without pre-commitment, the agent can
rationalize any observation post-hoc, weakening the
discrepancy signal that drives revision.
This ablation produces a notable decline, indicating that
pre-commitment is important not because it improves the individual
predictions, but because it ensures that each experiment generates
an unambiguous, falsifiable test of the current hypotheses.

\noindent\textbf{w/o Abduction (No Latent State Inference).}
Predictions are derived directly from the structural equations
without conditioning on accumulated evidence to infer latent
variable values. The core discovery loop remains intact, but
predictions become less calibrated because each hypothesis is not
grounded in what previous experiments revealed about the latent
state.
This ablation produces the mildest degradation among the three,
suggesting that abduction primarily refines prediction quality
rather than driving structural discovery.

\begin{table}[h]
\centering
\caption{\textbf{Ablation study on EvoSCM.}
We evaluate the effects of SCM representation, discriminative action,
structured revision, deductive validation, induction, pre-committed prediction,
and abductive inference.}
\label{tab:additional_ablation}

\renewcommand{\arraystretch}{1.2}
\setlength{\tabcolsep}{6pt}

\resizebox{\linewidth}{!}{%
\begin{tabular}{lcc}
\toprule
\textbf{Method}
& \textbf{$\langle$Explanation$\rangle$ $\uparrow$}
& \textbf{norm $\langle$MSE$\rangle$ $\downarrow$} \\
\midrule

Full EvoSCM
& \textbf{75.09\%}
& \textbf{2.77e-4} \\

w/o SCM (Free-Form Textual Hypotheses)
& 48.18\%
& 7.82e-2 \\

w/o Discriminative Action (Random Action)
& 47.45\%
& 1.05e-1 \\

w/o Structured Revision (Free-Form SCM Rewriting)
& 70.73\%
& 5.45e-4 \\ 

w/o Deduction (No Evidence Validation or Consistency Checking)
& 67.09\%
& 8.83e-4 \\

w/o Induction (No Correction Rules)
& 66.37\%
& 9.94e-4 \\

w/o Pre-committed Prediction (Only Observations)
& 63.47\%
& 4.79e-3 \\

w/o Abduction (No Latent State Inference)
& 69.58\%
& 5.74e-4 \\

\bottomrule
\end{tabular}%
}

\end{table}

\subsection{Additional Case Studies}

\begin{figure}[t]
    \centering
    \includegraphics[width=\linewidth]{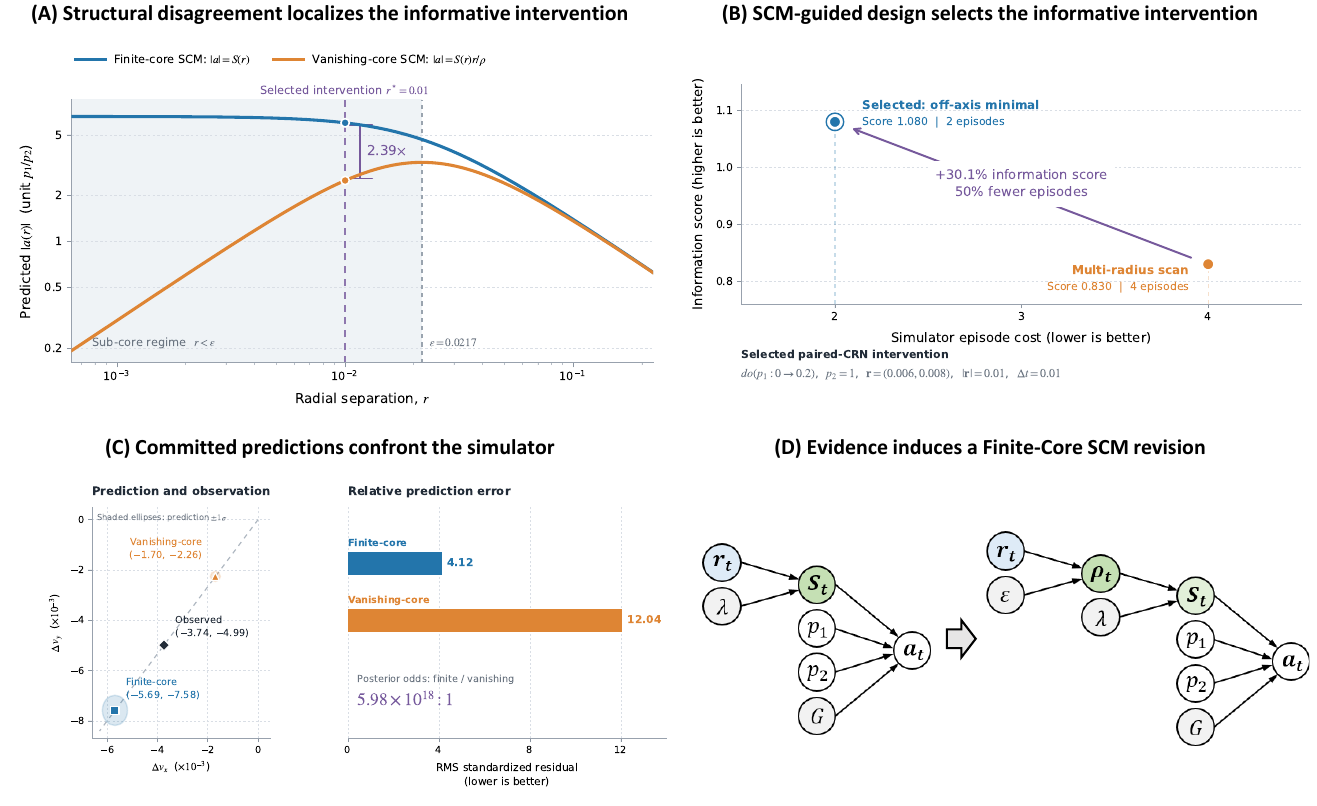}
    \caption{\textbf{Detailed case study: finite core recovery on
      the Yukawa world in
      DiscoverPhysics}~\citep{wiemann2026discoverphysics}.
      (A)~Surviving hypotheses diverge at short range,
      localizing a discriminative intervention at $r^* = 0.01$.
      (B)~EvoSCM selects the higher-information,
      lower-cost experiment.
      (C)~Pre-committed predictions favor the
      finite-core hypothesis.
      (D)~A localized revision inserts
      $r_t \to \rho_t \to S_t$ while preserving the rest of
      the causal structure.
    }
    \label{fig:analysis3}
\end{figure}

\paragraph{Detailed case study on the Yukawa world.}
Figure~\ref{fig:analysis3} zooms into a single revision event
to show how the four stages of EvoSCM's loop jointly recover the
finite core from Figure~\ref{fig:analysis2}.
(A)~After identifying the screened Yukawa mechanism, the surviving
hypotheses agree in the far field but diverge at short range:
predicted accelerations differ by $2.39\times$ at $r^* = 0.01$,
making this the maximally discriminative intervention
point.
(B)~EvoSCM selects a paired-CRN off-axis intervention over a
multi-radius scan for its higher information score and lower
cost ($1.080$ vs.\ $0.830$; 2 vs.\ 4 episodes).
(C)~The finite-core hypothesis produces a substantially smaller
residual than the vanishing-core alternative ($4.12\sigma$ vs.\
$12.04\sigma$), enabling decisive model selection.
(D)~The resulting revision inserts a softened distance
$\rho_t = \sqrt{r_t^2 + \varepsilon^2}$, replacing the direct
edge $r_t \to S_t$ with $r_t \to \rho_t \to S_t$ while
preserving the learned screening kernel and coupling
structure.

\begin{figure}[t]
    \centering
    \includegraphics[width=\linewidth]{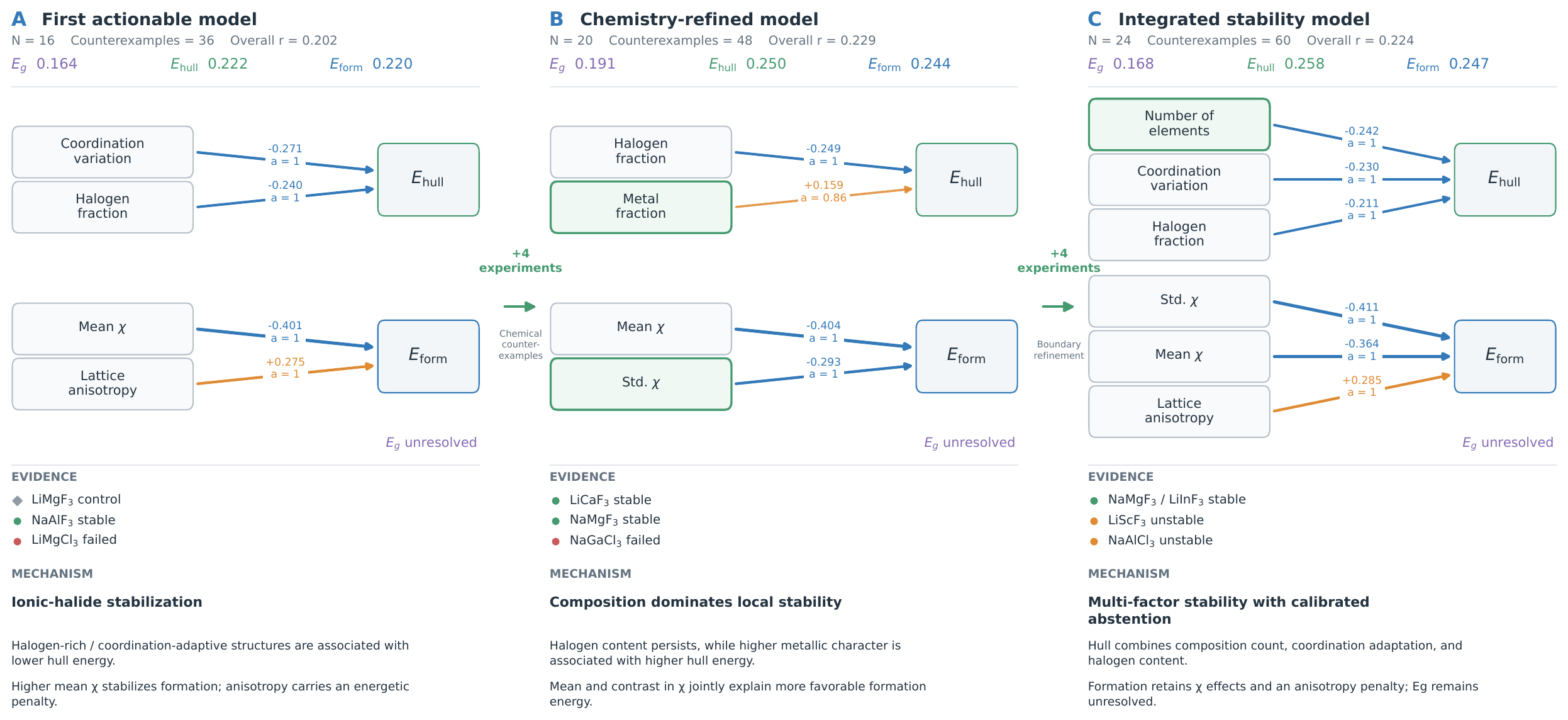}
    \caption{\textbf{SCM hypothesis evolution for solid-state electrolyte discovery on LLEMA}~\citep{abhyankar2026llema}. EvoSCM progressively refines its causal model of material stability across three stages of active experimentation. Edge color denotes the sign of the standardized coefficient, edge width its magnitude, opacity indicates sign agreement across the population of $K=7$ candidate SCMs, and green-outlined variables denote newly promoted mechanisms.}
    \label{fig:analysis4}
\end{figure}

\paragraph{Case Study on LLEMA.}
Figure~\ref{fig:analysis4} illustrates the evolution of EvoSCM's task-specific epistemic model for solid-state electrolyte discovery on LLEMA~\citep{abhyankar2026llema}.
At Stage~A, after 16 oracle-evaluated observations, EvoSCM forms its first actionable hypothesis: higher halogen content and coordination adaptability are associated with lower energy above the convex hull, while higher mean electronegativity favors formation energy and lattice anisotropy introduces an energetic penalty.
Four additional experiments, including informative chemical counterexamples, lead to Stage~B, where unsupported relations are revised and composition-related mechanisms are promoted: metallic fraction becomes positively associated with hull energy, while electronegativity contrast complements mean electronegativity in explaining formation energy.
After another four boundary-refinement experiments, Stage~C integrates elemental diversity, coordination variation, halogen content, electronegativity statistics, and lattice anisotropy into a multi-factor stability hypothesis.
The model continues to abstain from positing band-gap relations because their estimated reliability remains below the acceptance threshold, demonstrating calibrated epistemic restraint rather than forcing a complete explanation.
These structures should be interpreted as local, intervention-guiding epistemic relations learned within the frozen ALIGNN evaluation environment, not as recovery of universal causal laws for solid-state electrolytes.

\begin{figure}[t]
    \centering
    \includegraphics[width=\linewidth]{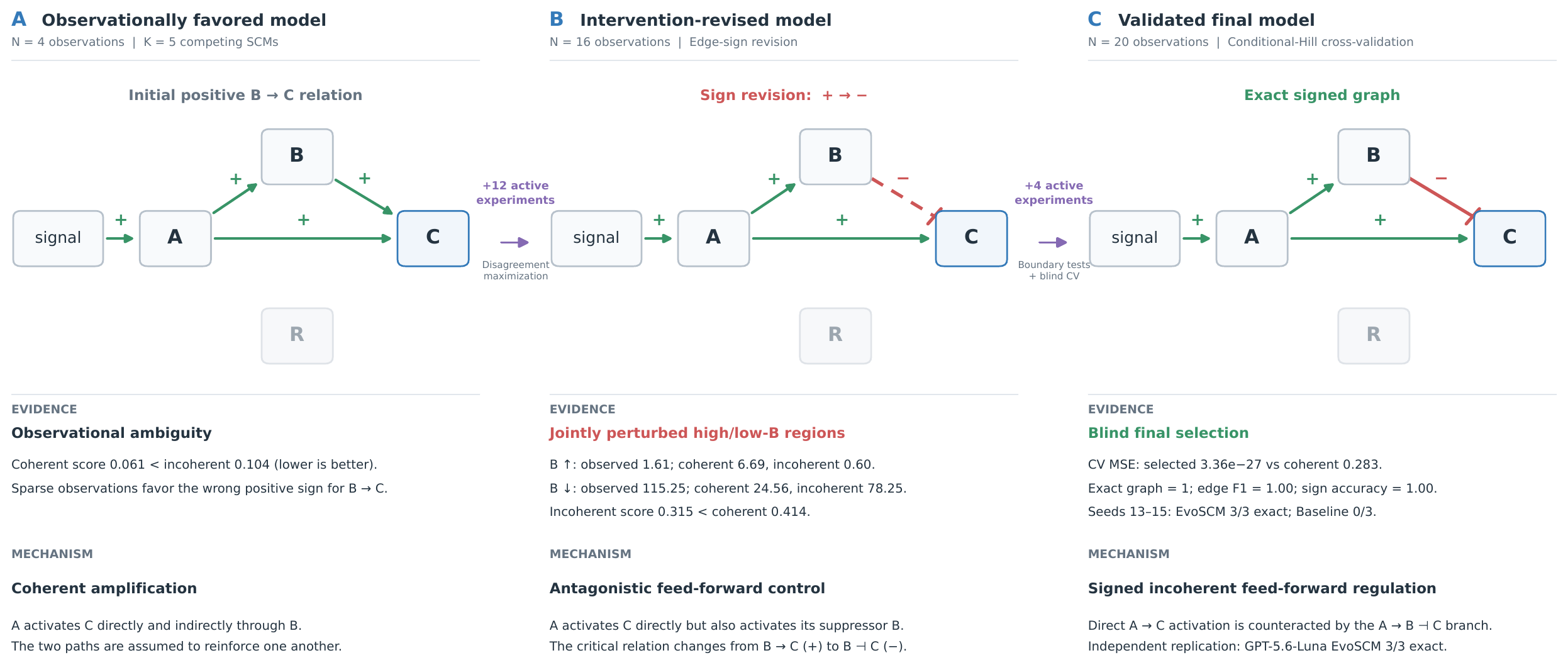}
    \caption{\textbf{SCM hypothesis evolution for gene regulatory network inference on ActiveSciBench-GRN}~\citep{kabra2026llm}. EvoSCM corrects an initial coherent feed-forward hypothesis to the ground-truth incoherent feed-forward motif through targeted soft interventions and sign revision.}
    \label{fig:analysis5}
\end{figure}

\paragraph{Case Study on ActiveSciBench-GRN.}
Figure~\ref{fig:analysis5} illustrates how EvoSCM revises a mechanistic hypothesis on ActiveSciBench-GRN~\citep{kabra2026llm}.
With only four initial observations, the coherent feed-forward model is preferred over the incoherent alternative (fit score $0.061$ vs.\ $0.104$, lower is better), leading to the incorrect hypothesis that $B$ activates $C$.
EvoSCM then selects continuous soft interventions in regions where the candidate SCMs disagree most strongly, jointly varying signal and perturbation multipliers rather than changing a single variable in isolation.
The resulting observations increasingly contradict the coherent explanation: for example, in two selected high- and low-$B$ regions, the observed reporter values are $1.61$ and $115.25$, compared with coherent-model predictions of $6.69$ and $24.56$ and incoherent-model predictions of $0.60$ and $78.25$, respectively.
After 16 observations, the working hypothesis revises the critical relation from $B\!\rightarrow\!C\,(+)$ to $B\!\dashv\!C\,(-)$, yielding an antagonistic feed-forward mechanism in which $A$ activates $C$ directly while also activating its suppressor $B$.
Following four additional boundary-testing experiments, the conditional-Hill cross-validation procedure selects the complete signed graph $\{\mathrm{signal}\!\rightarrow\!A,\ A\!\rightarrow\!B,\ A\!\rightarrow\!C,\ B\!\dashv\!C\}$ with a MSE of $3.36\times10^{-27}$, compared with $0.283$ for the coherent alternative.
Because coherent and incoherent motifs both belong to the public candidate grammar, this case demonstrates active discrimination, sign revision, and validation among competing biological mechanisms rather than unconstrained de novo discovery.

\section{Algorithmic Details}
\label{sec:algorithmic_details}

\subsection{SCM Representation}
\label{sec:scm_representation}

EvoSCM represents each SCM hypothesis
$\mathcal{H}_k^t = (\mathbf{V}, \mathbf{U}_k^t, G_k^t,
\mathbf{F}_k^t, \boldsymbol{\Theta}_k^t)$ as a structured
JSON object that the LLM can read, reason over, and edit.
The representation encodes three core components of the SCM
tuple alongside operational metadata.

\paragraph{Causal Graph.}
The graph $G_k^t$ is stored as an edge list of directed causal
relationships. Each variable is annotated with its role
(\texttt{state}, \texttt{parameter}) and observability, enabling
the Revision step to identify which variables can be added or
removed:
\begin{quote}
\small
\texttt{"variables": [} \\
\texttt{~~\{"name": "r\_t", "role": "state",
       "observed": true\},} \\
\texttt{~~\{"name": "C", "role": "parameter",
       "observed": false\}} \\
\texttt{],} \\
\texttt{"edges": [} \\
\texttt{~~\{"source": "r\_t", "target": "a\_t"\},} \\
\texttt{~~\{"source": "p1", "target": "a\_t"\}} \\
\texttt{]}
\end{quote}

\paragraph{Structural Equations.}
Each mechanism $f_i \in \mathbf{F}_k^t$ is stored as a symbolic
expression string together with its parent variables:
\begin{quote}
\small
\texttt{"mechanisms": [\{} \\
\texttt{~~"target": "a\_t",} \\
\texttt{~~"parents": ["r\_t", "p1", "p2"],} \\
\texttt{~~"equation": "a\_t = -C*(p1/p2)*r\_vec/}\\
\texttt{~~~~~~~~~~~~~(r**2 + eps**2)"} \\
\texttt{\}]}
\end{quote}
Storing the symbolic expression alongside its declared parents
allows the LLM to reason about functional dependencies and
supports direct numerical evaluation during the Prediction step.

\paragraph{Parameters.}
Parameters $\boldsymbol{\Theta}_k^t$ are stored with point
estimates and optional bounds:
\begin{quote}
\small
\texttt{"parameters": \{} \\
\texttt{~~"C": \{"estimate": 0.1585,} \\
\texttt{~~~~~~~~"lower": 0.155, "upper": 0.163\}} \\
\texttt{\}}
\end{quote}
Separating parameters from functional forms enables the
Update Parameter operator to adjust quantitative values
without modifying the structural equations.

\paragraph{Operational Metadata.}
Beyond the formal SCM tuple, each hypothesis carries metadata
that supports the discovery loop: a natural-language description
summarizing the causal explanation, a list of assumptions
(e.g., \texttt{"isotropic"}, \texttt{"static"}), falsification
conditions that would trigger revision, the hypothesis status
(\texttt{active} or \texttt{pruned}), and a revision counter
tracking how many times the hypothesis has been edited.
This metadata is not part of the formal SCM but provides context
that helps the LLM generate more targeted revisions and more
interpretable explanations.
A complete example of the JSON representation is provided in
Listing~\ref{lst:scm_example}.

\paragraph{Population Management.}
The population $\mathcal{P}_t = \{\mathcal{H}_1^t, \ldots,
\mathcal{H}_K^t\}$ is maintained as an indexed collection of
JSON objects. At each round, the serialized population and the
accumulated evidence $\mathcal{D}_t$ are provided to the LLM
so that inter-hypothesis comparisons (e.g., prediction
divergence for causal experiment design) can be
performed within the same call. In practice, the population
sizes used in our experiments ($K \leq 8$) remain well within
the context limits of all evaluated LLM backbones.

\begin{figure}[t]
\begin{lstlisting}[
  basicstyle=\ttfamily\scriptsize,
  frame=single,
  caption={Example SCM hypothesis representation
  (Yukawa environment).},
  label={lst:scm_example},
  breaklines=true
]
{
  "id": "H_screened_K1",
  "name": "Screened 2D Helmholtz field",
  "description": "Static isotropic screened field
                  with signed p1/p2 coupling.",
  "variables": [
    {"name": "pos2_t", "role": "state",
     "observed": true},
    {"name": "p1", "role": "source_control",
     "observed": true},
    {"name": "p2", "role": "response_inertia",
     "observed": true},
    {"name": "lambda", "role": "parameter",
     "observed": false}
  ],
  "edges": [
    {"source": "pos2_t",
     "target": "field_gradient_t"},
    {"source": "lambda",
     "target": "field_gradient_t"},
    {"source": "p1", "target": "a_t"},
    {"source": "p2", "target": "a_t"},
    {"source": "field_gradient_t",
     "target": "a_t"},
    {"source": "a_t",
     "target": "velocity2_next"}
  ],
  "mechanisms": [{
    "target": "a_t",
    "parents": ["pos2_t", "p1", "p2",
                "G", "lambda"],
    "equation": "a_t = -G*(p1/p2)*K1(r/lambda)/(2*pi*lambda)*pos2_t/r"
  }],
  "parameters": {
    "G": {"estimate": 0.9992,
          "lower": 0.05, "upper": 5.0},
    "lambda": {"estimate": 1.9999,
               "lower": 0.2, "upper": 10.0},
    "epsilon": {"estimate": 0.035,
                "lower": 1e-6, "upper": 0.2}
  },
  "assumptions": [
    "static",
    "isotropic",
    "velocity independent",
    "visible signs of p1 and p2 are causal"
  ],
  "falsifiers": [
    "far-range responses follow a constant
     power law rather than exponential
     K1 suppression"
  ],
  "status": "active",
  "revision": 6
}
\end{lstlisting}
\end{figure}

\subsection{Prompt Design}
\label{sec:prompt_design}

Each stage of the EvoSCM discovery loop is implemented as a
structured LLM call. All calls share a common preamble
containing the serialized hypothesis population
$\mathcal{P}_t$, the accumulated evidence $\mathcal{D}_t$,
and a description of the observable variables $\mathbf{V}$.
Below we summarize the stage-specific instructions; full
prompt templates are included in the released code.

\paragraph{Abduction.}
The LLM receives each hypothesis $\mathcal{H}_k^t$ with the
accumulated evidence and is instructed to infer latent
variable values consistent with the observed data:
\begin{quote}
\small\textit{Given the following SCM hypothesis and
experimental evidence, infer the values of the latent
exogenous variables that best explain the observed outcomes.
For each latent variable, provide a point estimate and a
brief justification based on the structural equations and
evidence.}
\end{quote}

\paragraph{Action.}
The LLM receives the full population $\mathcal{P}_t$ and is
instructed to select the intervention that maximally
discriminates between competing hypotheses:
\begin{quote}
\small\textit{You have $K$ competing SCM hypotheses. Design
an intervention $\mathrm{do}(\mathbf{X} = \mathbf{x})$ that
maximizes the disagreement among their predicted outcomes.
Identify the variable(s) to intervene on and the value(s) to
set, prioritizing interventions where different hypotheses
make the most divergent predictions.}
\end{quote}

\paragraph{Prediction.}
For each hypothesis $\mathcal{H}_k^t$, the LLM derives the
predicted outcome using that hypothesis's structural equations
and abduced latent states. The prompt enforces pre-commitment:
\begin{quote}
\small\textit{Using the structural equations and parameters of
this hypothesis, and the latent variable values inferred
during abduction, compute the predicted outcome of the
intervention $\mathrm{do}(\mathbf{X} = \mathbf{x})$. Commit
to a specific numerical prediction before observing the
experimental result.}
\end{quote}

\paragraph{Induction.}
The LLM receives each hypothesis's committed prediction
alongside the observed outcome and is instructed to identify
systematic patterns of failure:
\begin{quote}
\small\textit{Compare the predicted outcome with the
experimental observation. Identify any systematic discrepancy
and distill it into a concise correction rule describing the
empirical pattern that the current hypothesis fails to capture
(e.g., ``the relationship between X and Y follows an
inverse-square law rather than a linear law'').}
\end{quote}

\paragraph{Revision.}
The LLM receives the correction rules and is instructed to
apply targeted structural edits. The prompt explicitly
enumerates the available operators:
\begin{quote}
\small\textit{Based on the correction rule, revise the SCM
hypothesis using one or more of the following operators:
(1) Add/Remove Edge: insert or remove a directed edge;
(2) Add/Remove Latent: introduce or remove a latent variable;
(3) Update Mechanism: change the functional form of a
structural equation;
(4) Update Parameter: adjust the numerical value of an
existing parameter.
Apply the minimal set of operators sufficient to accommodate
the correction rule. Prefer shallow revisions (parameter and
mechanism updates) when possible, resorting to structural
edits only when the evidence warrants them.}
\end{quote}

\paragraph{Deduction.}
The LLM receives each revised hypothesis and performs two
validation checks:
\begin{quote}
\small\textit{Validate the revised hypothesis:
(1) Evidence Validation: derive predictions for all past
interventions in the accumulated evidence and verify that the
revised hypothesis accounts for previous observations without
introducing new contradictions.
(2) Consistency Check: verify that the causal graph is a valid
DAG, all structural equations are well-defined given the
revised graph, and the posited mechanisms are mutually
compatible.
Report whether the hypothesis passes both checks.}
\end{quote}

\subsection{Pseudocode}
\label{sec:pseudocode}

Algorithm~\ref{alg:evoscm} presents the complete EvoSCM
discovery loop. Each iteration comprises two phases: Causal
Experiment Design (Section~\ref{sec:experiment_design})
selects and executes a discriminative intervention, and SCM
Hypothesis Revision (Section~\ref{sec:revision}) transforms
the population through inductive correction, structured
revision, and deductive validation. After $T$ rounds, the
evolved population $\mathcal{P}_T$ is returned for downstream
inference (Section~\ref{sec:inference}).

\begin{algorithm}[t]
\caption{EvoSCM Discovery Loop}
\label{alg:evoscm}
\begin{algorithmic}[1]
\REQUIRE Environment $\mathcal{E}$, budget $T$,
         population size $K$, LLM backbone $\mathcal{L}$
\ENSURE Evolved population $\mathcal{P}_T$

\STATE Initialize $\mathcal{P}_0 = \{\mathcal{H}_1^0,
       \ldots, \mathcal{H}_K^0\}$
\STATE $\mathcal{D}_0 \leftarrow \emptyset$

\FOR{$t = 0$ \TO $T - 1$}

    \STATE \textcolor{gray}{\textit{\% Causal Experiment
           Design (\S\ref{sec:experiment_design})}}

    \FOR{$k = 1$ \TO $|\mathcal{P}_t|$}
        \STATE $\hat{\mathbf{U}}_k^t \leftarrow
               \textsc{Abduction}(\mathcal{H}_k^t,\,
               \mathcal{D}_t,\, \mathcal{L})$
    \ENDFOR

    \STATE $\mathbf{x}_{t+1}^* \leftarrow
           \textsc{Action}(\mathcal{P}_t,\,
           \mathcal{D}_t,\, \mathcal{L})$

    \FOR{$k = 1$ \TO $|\mathcal{P}_t|$}
        \STATE $\mathbf{y}_k^{\mathrm{pre}} \leftarrow
               \textsc{Prediction}(\mathcal{H}_k^t,\,
               \hat{\mathbf{U}}_k^t,\,
               \mathbf{x}_{t+1}^*,\, \mathcal{L})$
    \ENDFOR

    \STATE $\mathbf{y}_{t+1}^{\mathrm{obs}} \sim
           P^*(\mathbf{Y} \mid
           \mathrm{do}(\mathbf{X}\!=\!\mathbf{x}_{t+1}^*))$
    \STATE $\mathcal{D}_{t+1} \leftarrow \mathcal{D}_t
           \cup \{(\mathbf{x}_{t+1}^*,\,
           \mathbf{y}_{t+1}^{\mathrm{obs}})\}$

    \STATE \textcolor{gray}{\textit{\% SCM Hypothesis
           Revision (\S\ref{sec:revision})}}

    \STATE $\mathcal{P}_{t+1} \leftarrow \emptyset$

    \FOR{$k = 1$ \TO $|\mathcal{P}_t|$}
        \STATE $\mathcal{R}_k \leftarrow
               \textsc{Induction}(\mathcal{H}_k^t,\,
               \mathbf{y}_k^{\mathrm{pre}},\,
               \mathbf{y}_{t+1}^{\mathrm{obs}},\,
               \mathcal{D}_{t+1},\, \mathcal{L})$
        \STATE $\tilde{\mathcal{H}}_k^{t+1} \leftarrow
               \textsc{Revision}(\mathcal{H}_k^t,\,
               \mathcal{R}_k,\, \mathcal{L})$
        \STATE $v_k \leftarrow
               \textsc{Deduction}(
               \tilde{\mathcal{H}}_k^{t+1},\,
               \mathcal{D}_{t+1},\, \mathcal{L})$
        \IF{$v_k = \texttt{pass}$}
            \STATE $\mathcal{P}_{t+1} \leftarrow
                   \mathcal{P}_{t+1} \cup
                   \{\tilde{\mathcal{H}}_k^{t+1}\}$
        \ENDIF
    \ENDFOR

\ENDFOR

\RETURN $\mathcal{P}_T$

\end{algorithmic}
\end{algorithm}

\section{Benchmark Details}
\label{app:benchmark_details}

\subsection{DiscoverPhysics}
\label{app:benchmark_discoverphysics}

\paragraph{Task definition and composition.}
DiscoverPhysics~\citep{wiemann2026discoverphysics} is an interactive
benchmark for discovering hidden laws of motion in simulated worlds
whose physics deliberately departs from familiar physical assumptions.
The complete benchmark comprises 22 curated worlds, including
11 publicly released worlds and 11 private worlds.
Table~\ref{tab:discoverphysics_worlds} summarizes the public worlds,
which cover unusual force laws, hidden sources and particle types,
preferred spatial directions, and time-dependent interactions.

\paragraph{Experimental interface.}
A discovery session begins with initial trajectory observations and
an experimental protocol, without revealing the underlying force law
or hidden particle types.
Each experiment specifies controllable particles through their initial
positions, initial velocities, generalized charges, and requested
measurement times.
The simulator evolves these particles together with any existing
particles and returns the requested positions and velocities.
Observations are generated on demand rather than retrieved from a
fixed dataset.
The interface also supports fitting the free parameters of a proposed
law against previously collected trajectories.
Observation noise is configurable, whereas the held-out trajectories
used for prediction evaluation are noise-free.

\paragraph{Evaluation.}
The final submission consists of a natural-language explanation and
an executable Python implementation of the inferred dynamics.
Predictive accuracy is evaluated on held-out particle trajectories.
For a world $w$, let $\mathcal{I}_w$ denote the evaluated
particle--time pairs, and let $\mathbf{r}_{i}(t)$ and
$\widehat{\mathbf{r}}_{i}(t)$ denote the reference and predicted
positions, respectively.
The normalized trajectory error is
\begin{equation}
    \mathrm{nMSE}_{w}
    =
    \frac{1}{V_w \lvert\mathcal{I}_w\rvert}
    \sum_{(i,t)\in\mathcal{I}_w}
    \left\|
        \widehat{\mathbf{r}}_{i}(t)-\mathbf{r}_{i}(t)
    \right\|_2^2,
    \label{eq:discoverphysics_nmse}
\end{equation}
where $V_w$ is the benchmark's reference trajectory variance for that
world.
Normalization accounts for differences in the natural motion scales
of different worlds, and trajectory errors are aggregated
geometrically across worlds.

Explanation quality is evaluated separately by an LLM judge using an
expert-written, world-specific rubric, with scores reported on a
normalized scale.
Discovery success requires both sufficient predictive accuracy and
sufficient explanation quality.
The $\mathrm{pass}@k$ metric measures the expected fraction of worlds
with at least one successful discovery among $k$ independent attempts.

\begin{table}[t]
    \centering
    \small
    \renewcommand{\arraystretch}{1.12}
    \caption{
        \textbf{Public worlds on DiscoverPhysics}.
        The table summarizes the principal discovery targets of the
        11 publicly released worlds, rather than the full
        22-world benchmark.
    }
    \label{tab:discoverphysics_worlds}
    \begin{tabularx}{\linewidth}{
        @{}l>{\raggedright\arraybackslash}X@{}
    }
        \toprule
        \textbf{World} & \textbf{Principal discovery target} \\
        \midrule
        \texttt{gravity}
        & Two-dimensional attraction with force magnitude
          proportional to $1/r$. \\

        \texttt{yukawa}
        & A screened interaction with exponential suppression
          at large distances. \\

        \texttt{fractional}
        & An anomalous power-law interaction governed by
          a fractional operator. \\

        \texttt{circle}
        & A fractional interaction observed through a
          multi-particle ring configuration. \\

        \texttt{three\_species}
        & Hidden particle classes with distinct attractive
          and repulsive couplings. \\

        \texttt{dark\_matter}
        & Additional forces generated by unobserved sources. \\

        \texttt{ether}
        & A central interaction combined with a
          preferred-direction body force. \\

        \texttt{hubble}
        & A central interaction combined with an outward,
          position-dependent contribution. \\

        \texttt{oscillator}
        & A time-dependent coupling that periodically switches
          between attraction and repulsion. \\

        \texttt{extra\_dimensions}
        & A crossover from $1/r$ force scaling at large distances
          to $1/r^2$ scaling at short distances. \\

        \texttt{coulomb\_easy}
        & An attractive inverse-square central interaction. \\
        \bottomrule
    \end{tabularx}
\end{table}

\subsection{LLEMA}
\label{app:benchmark_llema}

\paragraph{Task definition and composition.}
LLEMA~\citep{abhyankar2026llema} introduces a suite of 14
application-driven materials-discovery tasks, publicly released as
LLEMABench.
Here, LLEMA refers to this benchmark suite, rather than the
evolutionary method introduced in the same work.
Each task specifies an application goal together with numerical
property constraints and, where applicable, compositional requirements.
The objective is to identify candidate materials satisfying these
requirements, rather than to recover a unique reference equation
or causal graph.
The released task specifications are summarized in
Table~\ref{tab:llema_tasks}.

\paragraph{Candidate representation and property evaluation.}
Candidates are represented by chemical compositions and
crystallographic information files (CIFs), which describe lattice
geometry, symmetry, and atomic sites.
The evaluation pipeline combines available Materials Project data
with pretrained property predictors, including CGCNN and ALIGNN,
to obtain task-relevant material properties.
These include electronic, mechanical, dielectric, and thermodynamic
quantities, with the required property set determined by the task.
Candidate validity is established by checking the resulting
properties and composition against the task specification.

The benchmark therefore evaluates computationally screened materials.
Its property estimates and stability assessments should not be
interpreted as direct experimental measurements or confirmation
of successful synthesis.

\paragraph{Evaluation.}
Hit rate measures the percentage of generated candidates satisfying
all task-specific constraints.
The stability metric measures the percentage that are both
task-valid and classified as thermodynamically stable.
For $M$ generated candidates, let $v_i\in\{0,1\}$ indicate whether
candidate $i$ satisfies the task constraints, and let
$s_i\in\{0,1\}$ indicate whether it passes the stability screen.
The two metrics, expressed as percentages, are
\begin{equation}
    \mathrm{H.R.}
    =
    \frac{100}{M}\sum_{i=1}^{M}v_i,
    \qquad
    \mathrm{Stab.}
    =
    \frac{100}{M}\sum_{i=1}^{M}v_i s_i.
    \label{eq:llema_metrics}
\end{equation}
Thus, the stability metric uses all generated candidates as its
denominator, rather than only the valid subset.
The original benchmark reports an energy-above-hull threshold of
$0.1$~eV/atom for thermodynamic stability screening.
It additionally uses Pareto-front analysis to assess trade-offs
among competing material properties.

\begin{table}[t]
    \centering
    \small
    \renewcommand{\arraystretch}{1.15}
    \caption{
        \textbf{Materials-discovery tasks on LLEMA.}
        Core property targets follow the publicly released task
        descriptions.
        The table summarizes task-level targets; thermodynamic
        stability is additionally assessed by the stability evaluator.
    }
    \label{tab:llema_tasks}
    \begin{tabularx}{\linewidth}{
        @{}
        >{\raggedright\arraybackslash}p{0.34\linewidth}
        >{\raggedright\arraybackslash}X
        @{}
    }
        \toprule
        \textbf{Task} & \textbf{Core property targets} \\
        \midrule
        Hard, stiff ceramics
        & $100\leq K\leq300$;
          $60\leq G\leq200$. \\

        Hard coating materials
        & $K\geq200$;
          $E_f\leq-1$;
          $E_g\geq3$. \\

        Structural materials for aerospace
        & $\rho\leq5$;
          $K\geq100$;
          $G\geq40$. \\

        High-$k$ dielectrics
        & $10\leq\kappa\leq90$;
          $2.5\leq E_g\leq6.5$. \\

        SAW/BAW acoustic substrates
        & $25\leq G\leq150$;
          $3.7\leq\kappa\leq95$. \\

        Solid-state electrolytes$^{\dagger}$
        & $E_f\leq-1$;
          $E_g\geq2$. \\

        Stable wide-bandgap semiconductors
        & $E_g\geq2.5$;
          $E_f\leq-1$. \\

        Photovoltaic absorbers$^{\ddagger}$
        & $0.7\leq E_g\leq2$;
          $E_f\leq0$. \\

        Piezo energy harvesters
        & $d\geq8$;
          $10\leq\kappa\leq8000$. \\

        Acousto-optic hybrids
        & $3\leq d\leq9$;
          $8\leq\kappa\leq85$. \\

        Electrically insulating dielectrics
        & $E_g\geq2.5$;
          $\kappa\geq8$. \\

        Transparent conductors
        & $E_g>3$;
          $50\leq\sigma\leq5000$. \\

        Low-density structural materials
        & $\rho\leq3.5$;
          $65\leq G\leq195$. \\

        Toxic-free perovskite oxides$^{\star}$
        & $E_g\geq2$;
          $90\leq K\leq135$. \\
        \bottomrule
    \end{tabularx}

    \par\smallskip
    \begin{minipage}{\linewidth}
        \footnotesize
        $K$ and $G$: bulk and shear moduli in GPa;
        $E_g$: band gap in eV;
        $E_f$: formation energy in eV/atom;
        $\rho$: density in $\mathrm{g\,cm^{-3}}$;
        $\kappa$: dimensionless dielectric constant;
        $d$: piezoelectric proxy in $\mathrm{pC\,N^{-1}}$;
        $\sigma$: electrical conductivity in
        $\mathrm{S\,cm^{-1}}$.
        For the two piezoelectric tasks, $d$ and $\kappa$ denote
        the corresponding proxies in the released specification.

        \smallskip
        $^{\dagger}$Requires at least one of Li, Na, K, Mg, Ca,
        or Al.
        $^{\ddagger}$Restricted to earth-abundant elements
        satisfying the benchmark's non-toxicity requirement.
        $^{\star}$Excludes Pb, Cd, Hg, Tl, Be, As, Sb, Se, U,
        and Th, and favors stable $\mathrm{ABO}_3$ oxides.
    \end{minipage}
\end{table}

\subsection{ActiveSciBench-GRN}
\label{app:benchmark_activescibench_grn}

\paragraph{Task definition and composition.}
ActiveSciBench-GRN, introduced with
LLM-AutoSciLab~\citep{kabra2026llm}, evaluates the recovery
of hidden gene-regulatory networks through active perturbation
experiments.
The reference suite comprises five regulatory motif families,
three benchmark versions, and three difficulty levels, yielding
$5\times3\times3=45$ task configurations per random seed.
Table~\ref{tab:grn_motifs} summarizes the motif families.
The discovery target is a signed directed graph specifying which
regulatory interactions exist, their directions, and whether they
are activating or repressing.
The inclusion of feedback motifs means that the target graphs
are not restricted to directed acyclic graphs.

\paragraph{Experimental interface.}
An experiment sets an upstream signal and multiplicative
perturbations of nodes $A$, $B$, $C$, and $R$.
Perturbation factors below or above one represent knockdown or
overexpression, respectively; one denotes the reference setting.
The public simulator returns a reporter response and a marker
panel for these nodes.
Responses are generated by motif-specific nonlinear regulatory
mechanisms rather than retrieved from a fixed observational dataset.

\paragraph{Difficulty levels.}
Difficulty is controlled through nonlinear response regimes.
Easy instances exhibit approximately linear responses;
medium instances introduce saturation that can obscure weak
regulatory effects; and hard instances emphasize stronger
nonlinearities and, for the relevant motifs, feedback-dependent
switching behavior.
The governing graph and dynamical parameters are withheld from
the learner.

\paragraph{Evaluation.}
Let
$\mathbf{A}^{\star},\widehat{\mathbf{A}}
\in\{-1,0,+1\}^{d\times d}$
denote the reference and predicted signed adjacency matrices,
where $+1$, $-1$, and $0$ represent activation, repression,
and absence of an edge, respectively.
Define the corresponding signed edge sets
$\mathcal{E}^{\star}$ and $\widehat{\mathcal{E}}$,
whose elements are
$(\text{source},\text{target},\text{sign})$ triples.
The public evaluator computes
\begin{equation}
    P
    =
    \frac{
        \lvert\widehat{\mathcal{E}}\cap\mathcal{E}^{\star}\rvert
    }{
        \lvert\widehat{\mathcal{E}}\rvert
    },
    \qquad
    R
    =
    \frac{
        \lvert\widehat{\mathcal{E}}\cap\mathcal{E}^{\star}\rvert
    }{
        \lvert\mathcal{E}^{\star}\rvert
    },
    \qquad
    F_1=\frac{2PR}{P+R},
    \label{eq:grn_edge_metrics}
\end{equation}
with undefined ratios assigned zero.
Thus, an edge is counted as correct only when its direction
and sign both match.

Sign accuracy measures sign agreement over directed edges
present in both graphs and is zero when this overlap is empty.
Exact graph accuracy requires equality of the complete signed
adjacency matrices.
The evaluator additionally provides motif accuracy, measuring
whether the predicted motif-family label is correct.

\begin{table}[t]
    \centering
    \small
    \renewcommand{\arraystretch}{1.15}
    \caption{
        \textbf{Regulatory motif families in ActiveSciBench-GRN.}
        Each family has three benchmark versions and three
        difficulty levels.
    }
    \label{tab:grn_motifs}
    \begin{tabularx}{\linewidth}{
        @{}
        >{\raggedright\arraybackslash}p{0.34\linewidth}
        >{\raggedright\arraybackslash}X
        @{}
    }
        \toprule
        \textbf{Motif family} & \textbf{Regulatory structure} \\
        \midrule
        Activation chain
        & A sequential activation cascade from an upstream
          signal to downstream regulators. \\

        Coherent feedforward loop
        & Direct and indirect activating paths with
          consistent regulatory effects. \\

        Incoherent feedforward loop
        & Competing paths combining activation
          and repression. \\

        Negative-feedback circuit
        & Downstream activation induces a repressive branch
          that limits the response. \\

        Toggle-switch circuit
        & Mutual repression supporting switching
          and state selection. \\
        \bottomrule
    \end{tabularx}
\end{table}

\section{Experiment Details}

\subsection{DiscoverPhysics}
\label{sec:discoverphysics_details}

\paragraph{Mapping SCM variables to physical quantities.}
The SCM maintained by EvoSCM is an \emph{epistemic} model of the
environment, not a copy of the simulator's ground-truth causal
program. For each hypothesis
$\mathcal{H}_k^t = (\mathbf{V}, \mathbf{U}_k^t, G_k^t,
\mathbf{F}_k^t, \boldsymbol{\Theta}_k^t)$, the variable set is
partitioned as
\[
\mathbf{V}_k^t
=
\mathbf{V}_{\mathrm{obs}}
\;\cup\;
\mathbf{V}_{\mathrm{act}}
\;\cup\;
\mathbf{V}_{\mathrm{lat},k}^t
\;\cup\;
\mathbf{V}_{\mathrm{der},k}^t.
\]
In two-particle environments,
$\mathbf{V}_{\mathrm{obs}}$ contains particle positions
$\mathbf{x}_t$, velocities $\mathbf{v}_t$, and the observed
trajectory; $\mathbf{V}_{\mathrm{act}}$ contains the public control
scalars $p_1$ and $p_2$, the initial position and velocity of the
mobile particle, and, when available, the absolute start time.
The hypothesis may introduce derived quantities
$\mathbf{V}_{\mathrm{der},k}^t$ such as the separation $r_t$,
radial direction $\hat{\mathbf{r}}_t$, field gradient
$\nabla\phi_t$, and acceleration $\mathbf{a}_t$.
The semantic roles of $p_1$ and $p_2$ are not hard-coded:
competing hypotheses may interpret them as source strength,
response inertia, charge, or nuisance controls, and these
interpretations are resolved through interventions.

The latent variables $\mathbf{V}_{\mathrm{lat},k}^t$ represent
unobserved physical causes proposed by the LLM, such as hidden
source locations, particle species, screening mechanisms, or
time-dependent couplings. Hypothesis-specific parameters
$\boldsymbol{\Theta}_k^t$ include quantities such as a coupling
constant $G$, radial exponent $q$, screening length $\lambda$,
and numerical regularization $\epsilon$.
These variables and parameters are inferred from observations;
they are not populated from the benchmark's hidden ground truth.

For population environments, the observed variables instead
contain positions and velocities of visible particles and probes.
Action variables include probe positions, velocities, and masses,
depending on the environment topology. Hidden dark sources and
latent species assignments may appear in the SCM but are not
directly manipulable by the agent.

\paragraph{Implementation of interventions.}
An intervention $\mathrm{do}(\mathbf{X} = \mathbf{x})$ is
implemented by assigning one or more fields of the public
experiment schema before the trajectory is generated. EvoSCM
cannot intervene on hidden physical laws, hidden simulator
parameters, or unexposed latent state.
Table~\ref{tab:discoverphysics_actions} summarizes the available
action spaces by environment topology.

\begin{table}[t]
\centering
\small
\caption{Public intervention variables in DiscoverPhysics.
  Measurement schedules determine when the trajectory is observed
  and are not treated as physical causes.}
\label{tab:discoverphysics_actions}
\begin{tabular}{@{}p{0.25\linewidth}p{0.67\linewidth}@{}}
\toprule
\textbf{Topology} & \textbf{Manipulable inputs} \\
\midrule
Two-particle
& $p_1$, $p_2$, initial position $\mathbf{x}_0$ and velocity
  $\mathbf{v}_0$ of the mobile particle, measurement
  schedule. \\[2pt]
Time-modulated
& Two-particle inputs plus the absolute
  \texttt{start\_time}. \\[2pt]
Circle
& Ring radius, initial tangential velocity, measurement
  schedule. \\[2pt]
Dark matter / three species
& Initial positions and velocities of neutral probes,
  measurement schedule. Hidden sources and species
  assignments cannot be set. \\[2pt]
Ether / Hubble
& Probe positions, velocities, masses, measurement
  schedule. \\
\bottomrule
\end{tabular}
\end{table}

EvoSCM supports both ordinary and paired-counterfactual
experiments. For an ordinary intervention, the selected input is
passed directly to the simulator. For a paired counterfactual,
the agent specifies a factual input $\mathbf{x}$ and a set of
sparse assignments $\{\delta_j\}$. The runtime evaluates
\[
\mathbf{y}^{(0)} = \mathrm{Sim}(\mathbf{x}),
\qquad
\mathbf{y}^{(j)} = \mathrm{Sim}
\bigl(\mathrm{do}_{\delta_j}(\mathbf{x})\bigr),
\]
and returns both branch outputs and paired differences
$\mathbf{y}^{(j)} - \mathbf{y}^{(0)}$. All branches share the
same hidden environment and a restored random-number state, so
paired differences use common random numbers and isolate the
declared intervention's effect. The factual rollout and each
counterfactual branch consume one simulator episode.

At each round, the LLM proposes two to four candidate experiment
designs together with pre-committed predictions from the
competing hypotheses. EvoSCM executes only the feasible design
with the largest posterior-weighted predictive disagreement,
ensuring that interventions are selected using predictions made
before observing the corresponding outcome.

\paragraph{Population size, rounds, and backbones.}
The maximum population size is $K_{\max} = 6$.
At the first round, the LLM initializes between three and six
structurally distinct hypotheses with uniform evidence weights,
irrespective of any confidence values reported by the LLM, to
avoid favoring a familiar textbook law before any experiment has
been performed.

Each session runs for at most $T_{\max} = 16$ LLM interaction
rounds and may terminate earlier when the agent submits its final
executable law. Since one counterfactual design can contain
multiple simulator branches, the number of LLM rounds and
simulator episodes are recorded separately. In paired
comparisons, EvoSCM uses the same world, noise seed, LLM
backbone, explanation judge, and maximum round count as the
matched baseline session. Its simulator-episode budget is
additionally capped by the number of episodes consumed by that
baseline session.

We evaluate with three LLM backbones: GPT-5.4~\citep{openai_gpt54},
GPT-5.5~\citep{openai_gpt55}, and GPT-5.6-Sol~\citep{openai_gpt56}. The
maximum output length is 8{,}192 tokens per call. Evaluation
results and reference explanations are not exposed to the
discovery agent.

\paragraph{Prompt construction.}
The system prompt is assembled from three components:
(i)~the benchmark-wide interactive discovery template,
(ii)~a topology-specific instruction file, and
(iii)~the EvoSCM protocol.
The topology-specific instructions are inserted into the master
template, after which the EvoSCM protocol is appended. At each
round, the runtime supplies a compact JSON state containing the
current hypothesis population, posterior weights, recent
prediction errors, the latest abduction, induction and deduction
records, completed intervention coverage, and remaining budget.
The LLM returns structured blocks: \texttt{<scm\_update>} for
abduction and hypothesis revision, \texttt{<design\_experiments>}
for candidate interventions and pre-committed predictions, and
\texttt{<final\_law>}, \texttt{<explanation>}, \texttt{<scm\_final>}
at termination. Complete prompt templates and schemas are included
in the released code.

\paragraph{Physics-specific adaptations.}
The EvoSCM loop is unchanged across environments, but the
following adaptations tailor evidence collection to physical-law
discovery.

\textit{Causal-invariance audits.}
For two-particle environments, the runtime tracks a set of
systematic audits rather than trusting the most familiar initial
law. These include a null-source intervention, controlled
magnitude changes in $p_2$, sign reversals of $p_1$ and $p_2$, a
multi-scale radius scan (at least four noise-cancelled responses
spanning at least an eight-fold radius range), and, when absolute
time is exposed, a phase sweep over \texttt{start\_time}. The
scale audit allows the agent to distinguish a constant power law
from Yukawa screening or a short-distance dimensional crossover.

\textit{Paired counterfactuals for noise cancellation.}
Paired counterfactuals cancel common observation noise and
estimate local causal responses. The runtime derives empirical
quantities (source and response-control exponents, local radial
slopes, sign relations, absolute-time responses) exclusively from
experiments observed by the agent. These quantities are returned
as evidence summaries; no hidden world label or ground-truth
equation is used.

\textit{Topology-specific action schemas.}
For population environments, EvoSCM reasons about latent source
populations through controlled probe placements, velocities, and
masses rather than intervening on hidden particles. For the circle
environment, experiments vary collective ring-level initial
conditions instead of individual two-particle controls.

\textit{Parameter calibration.}
Continuous parameters can be calibrated using the benchmark's
trajectory-level MSE-fitting tool on trajectories collected during
discovery. The final output is an executable Python law, and the
runtime checks its implied direction, control scaling, radial
dependence, and time dependence against observation-derived causal
evidence before accepting the submission.

\subsection{LLEMA}
\label{app:exp_llema}

\paragraph{Mapping SCM variables to material properties.}
For each task $\tau$, EvoSCM maintains an epistemic SCM over
\[
\mathbf{V}_\tau
= \mathbf{X}_{\mathrm{mat}} \cup \mathbf{Y}_\tau,
\]
where $\mathbf{X}_{\mathrm{mat}}$ contains composition and
structure descriptors and $\mathbf{Y}_\tau$ contains the target
properties required by the task. The causal direction is fixed
by domain knowledge:
\[
X_j \longrightarrow Y_{\tau,q},
\qquad
X_j \in \mathbf{X}_{\mathrm{mat}},
\quad
Y_{\tau,q} \in \mathbf{Y}_\tau.
\]
EvoSCM therefore does not search over arbitrary causal DAGs in
this domain. Instead, it evolves the active descriptor set, edge
signs and magnitudes, uncertainty estimates, and the relative
support for competing local mechanisms.

The descriptor variables include structural quantities (volume
per atom, density, lattice anisotropy, minimum pair distance,
nearest-neighbor statistics), compositional quantities
(composition entropy, elemental fractions, electronegativity
range), and atomic-level statistics (mean and standard deviation
of atomic number, electronegativity, mass, and radius). All
descriptors are deterministically computed from the generated
crystal using \textsc{Pymatgen}. For every task, energy above
hull ($E_{\mathrm{hull}}$) is included as an additional stability
outcome even when it is not a primary objective.
Table~\ref{tab:llema_scm_variables} shows the task-to-property
mapping.

\begin{table}[t]
\centering
\small
\caption{Mapping from LLEMA tasks to target property variables.}
\label{tab:llema_scm_variables}
\setlength{\tabcolsep}{4pt}
\begin{tabular}{p{0.34\linewidth}p{0.58\linewidth}}
\toprule
\textbf{Task} & \textbf{Outcomes in $\mathbf{Y}_\tau$} \\
\midrule
Hard, Stiff Ceramics
    & Bulk modulus; shear modulus \\
Hard Coating Materials
    & Bulk modulus; formation energy; band gap \\
Aerospace Materials
    & Density; bulk modulus; shear modulus \\
High-$k$ Dielectrics
    & Dielectric constant; band gap \\
SAW/BAW Acoustic Substrates
    & Shear modulus; dielectric constant \\
Solid-State Electrolytes
    & Formation energy; band gap \\
Wide-Bandgap Semiconductors
    & Band gap; formation energy \\
Photovoltaic Absorbers
    & Band gap; formation energy \\
Piezo Energy Harvesters
    & Piezoelectric coeff.; dielectric response \\
Acousto-optic Hybrids
    & Piezoelectric coeff.; dielectric response \\
Insulating Dielectrics
    & Band gap; dielectric constant \\
Transparent Conductors
    & Band gap; electrical conductivity \\
Low-Density Structures
    & Density; shear modulus \\
Perovskite Oxides
    & Band gap; bulk modulus \\
\bottomrule
\end{tabular}
\end{table}

\paragraph{SCM hypothesis population.}
We maintain $K = 7$ competing SCM hypotheses (not to be confused
with the number of material proposals). Each hypothesis is a
bootstrap sparse ridge mechanism fitted independently for each
property outcome. Features are standardized and screened per
property, with a ridge coefficient of 2.0. Bootstrap resampling
produces seven competing signed mechanisms; their median
coefficient, sign agreement, and coefficient variance define the
ensemble-level edge estimate and uncertainty.

The property-wise mechanisms are complemented by three additional
components: (i)~a calibrated LLM chemistry prior, (ii)~non-parametric
local mechanisms for the binary hit and stability outcomes, and
(iii)~a surrogate-support counterexample frontier. The numerical
SCM starts with a reliability prior of 0.15, while the LLM
chemistry prior is initialized at 0.25. Reliability is updated
using pre-committed prediction error and parent-child delta-sign
accuracy. Outcome-specific edges are withheld from the planner
when their calibrated reliability falls below 0.20, allowing the
system to retain a partially specified SCM rather than forcing
unsupported edges.

\paragraph{From material design to SCM interventions.}
EvoSCM does not directly apply interventions such as
$\mathrm{do}(\texttt{halogen\_fraction} = x)$, because modifying
a material generally changes several descriptors simultaneously.
Instead, the action variable is a chemically explicit
transformation
\[
a_t \colon m_{\mathrm{parent}} \longrightarrow m_{\mathrm{child}},
\]
where each proposed action records its type, changed factor,
source and target values, and parent formula, making the
intervention auditable. Supported transformations include:
\begin{itemize}[itemsep=2pt]
\item elemental substitutions (same-group, oxidation-state-preserving,
      or chemically similar);
\item stoichiometric and redox-aware composition changes;
\item crystal-prototype-preserving or prototype-changing
      interventions;
\item canonical polymorph realization at fixed composition;
\item isotropic lattice strain, polar displacement, and
      coordination changes at fixed composition;
\item stability-boundary, Pareto-frontier, and
      controlled-disagreement experiments.
\end{itemize}

A deterministic candidate compiler converts each intervention
intention into an evaluable crystal. The compiler can normalize
the formula, ground a composition in a common prototype, generate
bounded structural counterfactuals, and repair geometry errors.
It never queries the property oracle. Each compiled candidate
retains its source and all repair actions, so compiler
transformations remain distinct from oracle-derived evidence.

\paragraph{Population size, rounds, and budget.}
We use $T = 5$ adaptive rounds. A shared initialization requests
eight candidates from the LLM and evaluates four diverse
candidates. In each subsequent round, the LLM proposes
$K_{\mathrm{proposal}} = 8$ materials, the compiler constructs at
most $K_{\mathrm{compiled}} = 40$ pre-oracle candidates, and the
policy evaluates $B = 4$ candidates. Each task-seed session
therefore uses $B_0 + TB = 4 + 5 \times 4 = 24$ oracle episodes
and five adaptive LLM calls. Compiler-generated candidates that
are not selected do not consume oracle budget.

The active memory retains at most six success and six failure
exemplars in the prompt, while the fitted SCM uses all valid
observations. Five deterministic island labels provide provenance
and diversity tracking. The formula exclusion memory contains at
most 96 previously tested formulas.

\paragraph{LLM Backbones.}
The main comparison uses Qwen3.6-35B-A3B~\citep{qwen36_35b_a3b} with reasoning mode disabled, a maximum output length of 12{,}288 tokens, and strict JSON schema enforcement.
The transfer experiment uses GPT-5.6-Luna~\citep{openai_gpt56} with the same configuration: reasoning mode disabled, the same token limit, and the same budget.

\paragraph{Property oracle interface.}
After structural validation, each candidate is serialized as a CIF
and evaluated on the properties required by the current task plus
$E_{\mathrm{hull}}$. All methods use a single frozen ALIGNN oracle
version. Table~\ref{tab:llema_oracle} lists the property models.

\begin{table}[t]
\centering
\small
\caption{ALIGNN property models used for oracle evaluation.}
\label{tab:llema_oracle}
\setlength{\tabcolsep}{4pt}
\begin{tabular}{@{}ll@{}}
\toprule
\textbf{Property} & \textbf{Model identifier} \\
\midrule
Band gap
    & \texttt{jv\_optb88vdw\_bandgap\_alignn} \\
Formation energy
    & \texttt{jv\_formation\_energy\_peratom\_alignn} \\
Energy above hull
    & \texttt{jv\_ehull\_alignn} \\
Bulk modulus
    & \texttt{jv\_bulk\_modulus\_kv\_alignn} \\
Shear modulus
    & \texttt{jv\_shear\_modulus\_gv\_alignn} \\
Dielectric response
    & \texttt{jv\_epsx\_alignn} \\
Piezoelectric (max $d_{ij}$)
    & \texttt{jv\_dfpt\_piezo\_max\_dij\_alignn} \\
Piezoelectric (dielectric)
    & \texttt{jv\_dfpt\_piezo\_max\_dielectric\_alignn} \\
Density
    & Deterministic (from crystal geometry) \\
Conductivity
    & Derived: $\sigma = \mathrm{PF}/S^2$ (S/cm) \\
\bottomrule
\end{tabular}
\end{table}

A candidate is a \emph{hit} if it is structurally valid and
satisfies every task-specific constraint. It is \emph{stable} if
$E_{\mathrm{hull}} < 0.1~\mathrm{eV/atom}$, and a \emph{stable
hit} if both conditions hold. For Transparent Conductors and
Low-Density Structures, stability is included in the hit
definition per the task specification. H.R.\ is the fraction of
evaluated candidates that are hits; Stab.\ is the fraction that
are stable hits.

Oracle results include the property vector, failed constraints,
and cache provenance. Cache keys depend on the CIF bytes, property
model, and oracle version. Invalid structures receive no oracle
call, and every prediction is stored before its outcome is
observed.

\paragraph{Domain-specific adaptations.}
The EvoSCM loop is shared across all 14 tasks, but its generic
intervention language is mapped to materials-specific actions
through five frozen adapters.

\textit{1. Prototype grounding.}
Recognized compositions (AB, AB$_2$, ABX$_3$, A$_3$BX) are
grounded in common crystal prototypes (rocksalt, zincblende,
wurtzite, rutile, fluorite, perovskite, etc.). Piezoelectric
ABX$_3$ hypotheses use a bounded polar tetragonal realization
rather than centrosymmetric cubic.

\textit{2. Stability-aware portfolio allocation.}
The four evaluated slots per round are allocated to a
chemistry-anchor candidate, a joint-feasibility hypothesis, a
counterexample-repair or structural-counterfactual experiment,
and a stability/Pareto experiment.

\textit{3. Fixed-composition structural counterfactuals.}
Stable hits and near misses can undergo bounded isotropic strain,
prototype realization, coordination changes, or polar displacement
while keeping composition fixed.

\textit{4. Property-family adapters.}
Band-gap/dielectric tasks use a bounded strain ladder from a
stable near miss; acoustic tasks compare canonical realizations at
fixed composition; piezoelectric tasks prioritize polar-displacement
interventions; density/modulus tasks include explicit Pareto and
objective-boundary probes.

\textit{5. Calibration controls.}
A frozen library of textbook structures (e.g.,
LiMgF$_3$/LiCaF$_3$ for electrolyte calibration,
ZnO/AlN/LiNbO$_3$ for piezoelectric calibration, MgO/CaF$_2$ for
dielectric calibration) supplies positive, negative, and boundary
controls. These consume one oracle slot, are recorded as
calibration interventions, and are not presented as novel
discoveries.

All adapters are deterministic, frozen in the released code, and
operate before oracle evaluation. They may use public chemistry,
composition, geometry, task constraints, and previous observations,
but never an unobserved oracle outcome.

\subsection{ActiveSciBench-GRN}
\label{app:activescibench_grn}

\paragraph{Mapping SCM variables to genes.}
EvoSCM represents each epistemic hypothesis as a steady-state SCM
over the variable set
\[
\mathbf{V} = \{S, X_A, X_B, X_C, X_R\},
\]
where $S$ is an externally controlled signaling input and
$X_A, X_B, X_C, X_R$ are the steady-state expression levels of
four abstract genes. Gene $C$ is the reporter gene: the scalar
reporter intensity returned by the simulator is a scaled readout
of $X_C$. In addition to this reporter, the agent observes the
complete marker panel $(X_A, X_B, X_C, X_R)$ after each
experiment.

A candidate SCM contains a signed directed graph
$G_k = (\mathbf{V}, E_k, \sigma_k)$, where $\sigma_{ij} = +1$
denotes activation and $\sigma_{ij} = -1$ denotes repression.
Local mechanisms are represented using signed Hill-type functions.
For each endogenous gene,
\[
X_j = m_j \left[
    b_j + a_j
    \prod_{i \in \mathrm{Pa}_{G_k}(j)}
    H_{\sigma_{ij}}(X_i;\, K_{ij},\, n_{ij})
\right],
\qquad j \in \{A, B, C, R\},
\]
where $H_{+1}$ and $H_{-1}$ denote activating and repressing Hill
functions, respectively. Feedback and toggle motifs are
interpreted as equilibrium solutions of potentially cyclic
regulatory graphs rather than DAGs. The fitted parameters are
epistemic approximations used for prediction and model
discrimination; they are not assumed to reproduce the simulator's
hidden kinetic parameters.

\paragraph{Mapping interventions to gene perturbations.}
An experimental action is defined as
\[
\mathbf{u} = (s,\, m_A,\, m_B,\, m_C,\, m_R),
\]
with a continuous signal level $s \in [0.05, 10]$ and continuous
perturbation multipliers $m_j \in [0.1, 4]$. A multiplier
$m_j = 1$ denotes the unperturbed condition, $m_j < 1$ represents
knockdown, and $m_j > 1$ represents overexpression.
These are \emph{soft interventions}: the agent applies
$\mathrm{do}(m_j = c)$, which scales the production of gene $j$,
rather than directly fixing its expression through
$\mathrm{do}(X_j = c)$.

At each round, the intervention selector combines LLM-proposed
experiments with 512 log-uniformly sampled points from the legal
intervention space. It selects four experiments that maximize
predictive disagreement among the current SCM hypotheses, with a
diversity term that discourages redundant interventions. Each
hypothesis's reporter predictions are committed before the
selected experiments are submitted to the simulator.

\paragraph{Population size, rounds, and backbones.}
Each session begins with $N_0 = 4$ shared log-uniform initial
experiments. EvoSCM then runs $T = 4$ active-discovery rounds,
each proposing $K = 5$ structurally diverse candidate SCMs and up
to five candidate interventions. After numerical fitting and graph
deduplication, the best five distinct candidates form the working
population. Four new experiments are executed per round, giving a
total budget of $N_{\mathrm{exp}} = N_0 + T \times 4 = 20$.
One session therefore makes four LLM calls and twenty simulator
calls.

We evaluate two backbone configurations. The local-model
experiments use Qwen3.6-35B-A3B~\citep{qwen36_35b_a3b} with
reasoning mode disabled, temperature 0.7, top-$p = 0.8$,
top-$k = 20$, and a maximum output length of 4{,}096 tokens.
The API-model experiments use
GPT-5.6-Luna~\citep{openai_gpt56} with a maximum output length
of 4{,}096 tokens. Both configurations share the same $K$, $T$,
intervention bounds, initial-design size, and simulator budget.

The fixed working population size $K = 5$ is distinct from the
candidate archive maintained by the final selector. The archive
retains every valid unique graph produced during the four rounds
and therefore has a data-dependent size.

\paragraph{Relation between the epistemic SCM and the target
network.}
The target regulatory network $G^*$ and each epistemic candidate
$G_k$ share the same labeled node set and signed-edge
representation. A candidate edge $A \xrightarrow{+} B$ directly
expresses the hypothesis that $A$ activates $B$, while
$R \xrightarrow{-} C$ hypothesizes repression of the reporter
gene. The predicted graph can therefore be compared directly
against the hidden signed adjacency matrix of $G^*$.
The agent receives a public universal hypothesis grammar
describing all five possible motif families, together with the
intervention interface and graph constraints. It is not given the
current session's motif identity, ground-truth edges, kinetic
parameters, evaluator scores, or task identifier. Candidate
graphs may contain at most six signed edges, cannot contain
self-loops, cannot use $S$ as a destination, and must include a
directed path from $S$ to $C$. The same grammar is supplied to
EvoSCM and the baseline.
This experiment therefore evaluates active discrimination,
revision, and selection within a public, biologically motivated
hypothesis class; it is not unconstrained de novo topology
discovery.

\paragraph{GRN-specific final selection.}
After the fourth round, EvoSCM retains all valid unique graph
candidates encountered during the session. For each archived
graph, it fits node-conditional signed Hill mechanisms using only
the observed interventions and marker values. Candidates are
ranked by five-fold grouped cross-validation using the mean
held-out squared log error over $\{A, B, C, R\}$. Ties are
resolved first by preferring the graph with fewer edges, then by
a deterministic canonical graph ordering.
This selector uses neither the target graph nor evaluator metrics,
and makes no additional LLM calls or simulator experiments.
The selected graph is frozen before the evaluator is invoked and
is used unchanged for all graph-recovery metrics. Exact Graph
Accuracy requires equality of the complete signed adjacency
matrix; Edge F1 is computed over signed edge triples; and Sign
Accuracy measures sign agreement on edges present in both the
prediction and target.

\section{Yukawa World Details}
\label{sec:yukawa-world-details}

This section relates the ground-truth law used by the Yukawa world to the
standard continuum Yukawa theory and clarifies the three stages of the SCM
evolution shown in Figure~\ref{fig:analysis2}.  The apparent difference from
the commonly quoted Yukawa potential, $\exp(-r/\lambda)/r$, is caused by
dimensionality: that expression is the Green function in three spatial
dimensions, whereas the DiscoverPhysics Yukawa world is two-dimensional.

\paragraph{Continuum field equation.}
Let $\mathbf{x}_1$ be the fixed source position, let
$\mathbf{r}=\mathbf{x}_2-\mathbf{x}_1$, $r=\lVert\mathbf{r}\rVert$, and
$\widehat{\mathbf{r}}=\mathbf{r}/r$.  With $m=\lambda^{-1}$, the static
screened field is governed by the modified Helmholtz equation
\begin{equation}
    (\nabla^2-m^2)\phi(\mathbf{x})
    = Gp_1\,\delta^{(d)}(\mathbf{x}-\mathbf{x}_1),
    \label{eq:yukawa-modified-helmholtz}
\end{equation}
where $G$ is the field coupling, $p_1$ is the source strength, and $d$ is the
number of spatial dimensions.  Fourier transformation gives
\begin{equation}
    \widetilde{\phi}(\mathbf{k})
    = -\frac{Gp_1}{\mathbf{k}^2+m^2}.
\end{equation}
Consequently, the $d$-dimensional Green function is
\begin{align}
    \mathcal{G}_d(r)
    &\equiv
    \int\frac{\mathrm{d}^d k}{(2\pi)^d}
    \frac{e^{i\mathbf{k}\cdot\mathbf{r}}}{\mathbf{k}^2+m^2} \\
    &=
    \frac{1}{(2\pi)^{d/2}}
    \left(\frac{m}{r}\right)^{d/2-1}
    K_{d/2-1}(mr),
    \label{eq:yukawa-green-general}
\end{align}
where $K_\nu$ denotes the modified Bessel function of the second kind.  The
continuum potential is $\phi_d(r)=-Gp_1\mathcal{G}_d(r)$.

For $d=3$, using
$K_{1/2}(z)=\sqrt{\pi/(2z)}e^{-z}$ recovers the familiar expression
\begin{equation}
    \phi_3(r)=-\frac{Gp_1}{4\pi}\frac{e^{-r/\lambda}}{r}.
    \label{eq:yukawa-3d}
\end{equation}
Thus, the usual exponentially screened inverse-distance potential is the
three-dimensional member of the same family, rather than the appropriate
ground truth for the present two-dimensional world.

\paragraph{Two-dimensional Yukawa potential and force.}
For $d=2$, Equation~\ref{eq:yukawa-green-general} instead gives
\begin{equation}
    \mathcal{G}_2(r)=\frac{1}{2\pi}K_0(mr),
    \qquad
    \phi_2(r)=-\frac{Gp_1}{2\pi}K_0(r/\lambda).
    \label{eq:yukawa-2d-potential}
\end{equation}
The mobile particle has response/inertia parameter $p_2$ and obeys
\begin{equation}
    \dot{\mathbf{x}}_2=\mathbf{v}_2,
    \qquad
    \dot{\mathbf{v}}_2
    =\mathbf{a}(\mathbf{r})
    =-\frac{1}{p_2}\nabla\phi_2(r).
    \label{eq:yukawa-continuous-dynamics}
\end{equation}
Using $\mathrm{d}K_0(z)/\mathrm{d}z=-K_1(z)$ yields
\begin{equation}
    \boxed{
    \mathbf{a}(\mathbf{r})
    =-G\frac{p_1}{p_2}
    \frac{K_1(r/\lambda)}{2\pi\lambda}
    \widehat{\mathbf{r}}
    }.
    \label{eq:yukawa-2d-acceleration}
\end{equation}
Equation~\ref{eq:yukawa-2d-acceleration} is therefore not an unrelated
discrete empirical law: it is the radial force obtained directly by taking
the gradient of the continuum two-dimensional Yukawa potential.

\paragraph{Operational ground truth in the simulator.}
The N-body backend regularizes close encounters with the softened radius
\begin{equation}
    \rho=\sqrt{r^2+\epsilon^2}.
    \label{eq:yukawa-softened-radius}
\end{equation}
It evaluates the radial kernel at $\rho$ while retaining the physical radial
direction $\widehat{\mathbf{r}}$.  The operational law that generates the
observed trajectories is therefore
\begin{equation}
    \boxed{
    \mathbf{a}_{\mathrm{sim}}(\mathbf{r})
    =-G\frac{p_1}{p_2}
    \frac{K_1(\rho/\lambda)}{2\pi\lambda}
    \widehat{\mathbf{r}},
    \qquad
    \rho=\sqrt{r^2+\epsilon^2}
    }.
    \label{eq:yukawa-operational-ground-truth}
\end{equation}
The benchmark parameters are
$G=1$, $\lambda=2$, and $\epsilon=0.05$.  The replacement $r\mapsto\rho$ is
a simulator-level numerical regularization, not an additional prediction of
the unregularized continuum Helmholtz equation.  In particular,
Equation~\ref{eq:yukawa-operational-ground-truth} approaches the continuum
law in Equation~\ref{eq:yukawa-2d-acceleration} when $r\gg\epsilon$, while
its magnitude remains finite when $r\lesssim\epsilon$.  The direction floor
$\delta=10^{-6}$ used in the learned executable law is likewise a numerical
safeguard of the agent implementation and is not a physical ground-truth
parameter.

The simulator advances the continuous state
$(\mathbf{x}_t,\mathbf{v}_t)$ using a fourth-order Yoshida symplectic
integrator with step size $\Delta t=0.005$.  Hence, the time-unrolled arrows
$\mathbf{a}_t\rightarrow\mathbf{v}_{t+1}\rightarrow\mathbf{r}_{t+1}$ in
Figure~\ref{fig:analysis2} represent the numerical state-transition layer,
whereas Equation~\ref{eq:yukawa-operational-ground-truth} is the scientific
mechanism recovered by the epistemic SCM.

\paragraph{Interpretation of the SCM evolution.}
The three stages in Figure~\ref{fig:analysis2} follow directly from distinct
asymptotic regimes of the same operational law.

First, for $\epsilon\ll r\ll\lambda$,
\begin{equation}
    K_1(z)=\frac{1}{z}+\mathcal{O}(z\log z),
\end{equation}
and therefore
\begin{equation}
    \mathbf{a}_{\mathrm{sim}}(\mathbf{r})
    \simeq
    -G\frac{p_1}{p_2}\frac{1}{2\pi r}
    \widehat{\mathbf{r}}.
    \label{eq:yukawa-near-asymptotic}
\end{equation}
The initial unscreened inverse-radius SCM is thus a valid local approximation
to the two-dimensional Yukawa force, rather than an arbitrary incorrect
prior.

Second, for $r\gg\lambda$,
\begin{equation}
    K_1(r/\lambda)
    \simeq
    \sqrt{\frac{\pi\lambda}{2r}}e^{-r/\lambda},
    \label{eq:yukawa-far-asymptotic}
\end{equation}
so a sufficiently broad radius sweep reveals the exponentially suppressed
tail.  This evidence motivates the SCM revision that introduces the screening
length $\lambda$ and the $K_1$ mechanism.

Finally, for $r\lesssim\epsilon$,
$\rho\simeq\epsilon$, and the force magnitude ceases to follow the divergent
$1/r$ near-field approximation.  Targeted sub-core interventions therefore
identify the additional causal path
\begin{equation}
    (r,\epsilon)\longrightarrow\rho
    \longrightarrow K_1(\rho/\lambda)
    \longrightarrow\mathbf{a},
\end{equation}
which explains the finite-core revision in the final SCM.  This last step is
a discovery of the simulator's operational regularization, whereas the
preceding $K_0$/$K_1$ structure is the recovery of the underlying continuum
two-dimensional Yukawa mechanism.  Accordingly, the final model in
Figure~\ref{fig:analysis2} is ground-truth-equivalent for the evaluated
N-body environment while retaining a clear correspondence to the continuum
Yukawa field theory.

\section{Additional Related Work}

\subsection{Self-Evolving LLM Agents}

Self-evolving LLM agents can improve from experience without updating their parameters~\citep{wang2024survey,tao2024survey,fang2025comprehensive}, spanning approaches such as prompt optimization, memory distillation, skill acquisition, workflow search, and multi-agent collaboration.

\paragraph{Prompt and Strategy Optimization.}
OPRO~\citep{yang2024large} frames prompt optimization as a black-box optimization problem, using LLMs to iteratively generate and refine prompts based on task performance.
PromptBreeder~\citep{fernando2024promptbreeder} applies an evolutionary algorithm to co-evolve task prompts and mutation operators, demonstrating that self-referential improvement can outperform hand-crafted prompt engineering.
EvoPrompt~\citep{guo2024connecting} similarly evolves discrete prompts using genetic algorithms and differential evolution, connecting prompt optimization to classical evolutionary computation.
These methods optimize the instructions given to the LLM but do not modify the agent's model of the external world, which is the focus of EvoSCM.

\paragraph{Reflective and Experiential Learning.}
Reflexion~\citep{shinn2023reflexion} introduces verbal reinforcement learning, where agents generate natural-language reflections on failures and store them in an episodic memory buffer to avoid repeating mistakes.
ExpeL~\citep{zhao2024expel} extracts transferable experiential insights from successful and failed trajectories, building a growing library of lessons that generalize across tasks.
Self-Refine~\citep{madaan2023self} demonstrates that LLMs can iteratively critique and improve their own outputs through multi-round feedback, without any external training signal.
These systems accumulate experiential knowledge that improves decision-making, but the accumulated knowledge takes the form of verbal reflections, heuristic rules, or trajectory summaries rather than executable scientific models with falsifiable predictions.

\paragraph{Multi-Agent and Workflow Evolution.}
MetaGPT~\citep{hong2024metagpt} organizes multiple LLM agents into structured workflows with role-based specialization, demonstrating that procedural organization improves complex task performance.
AutoGen~\citep{wu2023autogen} provides a general-purpose framework for multi-agent conversations with customizable interaction patterns.
AgentVerse~\citep{chen2023agentverse} explores how groups of agents can collaboratively evolve their problem-solving strategies through dynamic team composition.
These approaches evolve the agent's organizational structure and communication protocols, a form of procedural self-evolution that is complementary to EvoSCM's epistemic self-evolution.

\subsection{Scientific Agents}

A separate body of work applies LLM agents to scientific research~\citep{ai4science2023impact,wei2025ai,ren2025towards}, spanning hypothesis generation, experiment design, result interpretation, and domain-specific applications across physics, chemistry, and biology.

\paragraph{Laboratory and Experimental Agents.}
ChemCrow~\citep{m2024augmenting} augments LLMs with chemistry-specific tools for synthesis planning, safety assessment, and property prediction, demonstrating end-to-end chemical research assistance.
SciAgents~\citep{ghafarollahi2025sciagents} employs multi-agent collaboration with graph-based reasoning for materials science discovery.
Agent Laboratory~\citep{schmidgall2025agent} automates the full research workflow from literature review through experimentation to report writing, using LLM-driven planning to coordinate each stage.
These systems demonstrate that LLM agents can interact with real or simulated scientific environments, but they typically do not maintain an explicit, revisable model of the underlying mechanisms that can be tested against experimental outcomes.

\paragraph{Hypothesis Generation and Evolution.}
SciMON~\citep{wang2024scimon} generates novel scientific ideas by mining literature for inspiration and iteratively refining hypotheses against existing work.
ResearchAgent~\citep{baek2025researchagent} generates research ideas by building entity-relation graphs from academic papers and using them to identify knowledge gaps.
FunSearch~\citep{romera2024mathematical} evolves programs through LLM-guided search to discover novel mathematical and scientific solutions, demonstrating that iterative LLM-based refinement can yield genuine discoveries.
While these systems contribute to automated hypothesis generation, they generally represent hypotheses in natural language rather than as executable causal models, limiting their ability to derive interventional predictions or trace prediction failures to specific structural commitments.

\end{document}